\documentclass{article}

\usepackage[final]{neurips_2024}

\usepackage[utf8]{inputenc} 
\usepackage[T1]{fontenc}    
\usepackage[hidelinks]{hyperref}      
\usepackage{url}            
\usepackage{booktabs}       
\usepackage{amsfonts}       
\usepackage{nicefrac}       
\usepackage{microtype}      
\usepackage{xcolor}         

\usepackage{amsmath}
\usepackage{amssymb}
\usepackage{mathtools}
\usepackage{amsthm}

\usepackage{multirow}
\usepackage{siunitx}
\usepackage{wrapfig}
\usepackage{dblfloatfix}
\usepackage{caption}
\usepackage{float}

\def\SPSB#1#2{\rlap{\textsuperscript{{#1}}}\SB{#2}}

\def\SB#1{\textsubscript{{#1}}}

\title{SOI: Scaling Down Computational Complexity by Estimating Partial States of the Model}

\author{%
  Grzegorz Stefański\textsuperscript{1}, Paweł Daniluk\textsuperscript{2}, Artur Szumaczuk\textsuperscript{2}, Jakub Tkaczuk\textsuperscript{2} \\
  \textsuperscript{1}Samsung AI Center Warsaw\quad\textsuperscript{2}Samsung R\&D Institute Poland\\
  \texttt{\{g.stefanski, p.daniluk, a.szumaczuk, j.tkaczuk\}@samsung.com} \\
}

\begin{document}

\maketitle

\begin{abstract}
  Consumer electronics used to follow the miniaturization trend described by Moore’s Law. Despite increased processing power in Microcontroller Units (MCUs), MCUs used in the smallest appliances are still not capable of running even moderately big, state-of-the-art artificial neural networks (ANNs) especially in time-sensitive scenarios. In this work, we present a novel method called Scattered Online Inference (SOI) that aims to reduce the computational complexity of ANNs. SOI leverages the continuity and seasonality of time-series data and model predictions, enabling extrapolation for processing speed improvements, particularly in deeper layers. By applying compression, SOI generates more general inner partial states of ANN, allowing skipping full model recalculation at each inference.
\end{abstract}

\section{Introduction}
Moore’s Law \citep{moore1998} predicts that the number of transistors on a microchip will double every 2 years. As the number of transistors increases, the computational capacity of systems-on-chip (SoCs) also grows. Accompanied by the further miniaturization of components, the increased complexity of SoCs allows for the production of more compact appliances with equal or higher capabilities. 

Despite the continuous development of more advanced hardware, the smallest appliances still cannot fully benefit from increasingly popular neural-based solutions, as they are not able to run them efficiently. Examples of such appliances include wireless in-ear headphones, smart watches, and AR glasses. Furthermore, the size of state-of-the-art neural networks is growing at  much faster rate than that described by Moore’s Law \citep{Xu2018}. Adding to the challenge, according to researchers \citep{Leiserson2020}, Moore’s Law is starting to decelerate due to the physical constraints of semiconductors, and the ``room at the bottom'' is depleting.

Currently available neural network technologies enable machines to outperform humans in numerous applications in terms of measured performance \citep{Nguyen2020, Bakhtin2022, Schrittwieser2020}. However, despite this achievement, the same models fall significantly short when it comes to energy efficiency compared to humans. The human brain consumes a mere 20 Watts of power \citep{Laughlin1998, Sengupta2014, Balasubramanian2021} and according to \citet{Xu2018} it is estimated to be over five orders of magnitude more energy efficient than modern neural networks.

This disparity in energy efficiency can be attributed to the common pursuit of the highest model quality in the literature, showcasing the full capabilities of the developed technology, often at the expense of efficiency. This behavior is justifiable due to the ease of comparing different solutions using well-defined metrics that are independent of the hardware and software. Conversely, estimating a model's energy efficiency is a more complex task, influenced by various factors including the running environment.

However, this trend within our community may prove restrictive, particularly for applications like real-time systems. These applications naturally demand optimal performance alongside high-efficiency solutions, rendering most current Deep Neural Networks (DNNs) impractical for such tasks. Furthermore, due to the substantial discrepancy between assumptions for high-efficiency on-device solutions and high-performing monolithic models, compressing these models may not be a viable means of applying state-of-the-art DNNs to real-time tasks.

The importance of neural system efficiency is also increasingly significant from ecological and economic standpoints \citep{lacoste2019,patterson2021}.

\subsection{Related Works}
The Short-Term Memory Convolution (STMC) \citep{Stefanski2023} was devised to enhance the efficiency of Convolutional Neural Networks (CNNs) inference by reducing computation requirements at each step via eliminating the need to recompute prior states. The authors achieved a notable 5.68-fold reduction in inference time and a 30\% decrease in memory consumption compared to the Incremental Inference method \citep{Romaniuk2020}. STMC enables the conversion of a standard CNN model, which typically requires an input of size at least as large as its receptive field, into a model capable of processing a single input frame at a time, akin to Long-Short Term Memory networks (LSTM). The SOI method is built upon the STMC foundation, offering distinct treatment of strided layers and yielding a compelling new \emph{inference pattern}\footnote{By the term of an inference pattern we mean a full computational graph of a single inference or of repeating sequence of inferences if they influence the computational graphs of the following model executions as in case of SOI or STMC.}. 

Routing methods constitute a popular category of algorithms tailored to optimize the inference process of Recurrent Neural Networks (RNNs). In the field of Natural Language Processing (NLP), \citet{yu2017} introduced an approach that involves traversing segments of the computational graph. This traversal is guided by decisions made by a reinforcement learning model following the processing of a fixed number of words. Another notable contribution by \citet{Campos2018} yielded an algorithm capable of selectively bypassing partial state updates within an RNN during inference, influenced by the input's length. In NLP terms, this concept can be compared to the mechanism of skipping words.

The research by \citet{jernite2017} introduced a distinct method to regulate computation within a recurrent unit. This method relied on a scheduler unit that facilitated partial updates to the network's state, only when the computational budget limit was reached. Meanwhile, \citet{seo2018} proposed an approach referred to as ``word skimming''. In this approach, the authors designed both small and large RNN models that could be interchangeably utilized for inference through the utilization of Gumbel softmax. The exploration of hybrid techniques combining jumping, skimming, and skipping was further advanced by \citet{hansen2018}, who published additional solutions in this direction.

The RNN routing methods have found successful applications in CNNs as well. \citet{Wang2018} introduced an approach that enables a model to learn a policy for skipping entire convolutional layers on a per-input basis. A similar concept, involving adaptive inference graphs conditioned on image inputs, was put forth by \citet{Veit2018}. Additionally, several authors have contributed methods for early-exit during CNN inference \citep{Bolukbasi2017, Teerapittayanon2016, Huang2017}. In these methods, the network is trained to skip portions of the computational graph towards the final stages, based on the characteristics of the input.

Other commonly employed methods for model optimization include pruning \citep{LeCun1989} and quantization \citep{Gray1998}. Importantly, both of these methods are not mutually exclusive and can coexist alongside routing methods within a single neural network.

\subsection{Novelty}

In this study, we introduce a method for reducing the computational cost of a ANN model while incurring only a negligible decrease in the model's performance. Importantly, these reductions are achieved with minimal alterations to the architecture, making it suitable for various tasks where factors such as energy consumption or time are of paramount importance.

Our approach involves the conversion of a conventional ANN model, initially trained to process segments of time-series data with arbitrary lengths in an offline mode, into a model that processes the data element by element, enabling real-time usage. Notably, our method builds upon the STMC technique 
\citep{Stefanski2023}. STMC is designed to perform each distinct operation exactly once. SOI extends this concept by omitting some operations
in a structured manner.

\begin{figure*}[b]
    \centering
    \ifdefined\enableColors
        \includegraphics[width=0.85\textwidth]{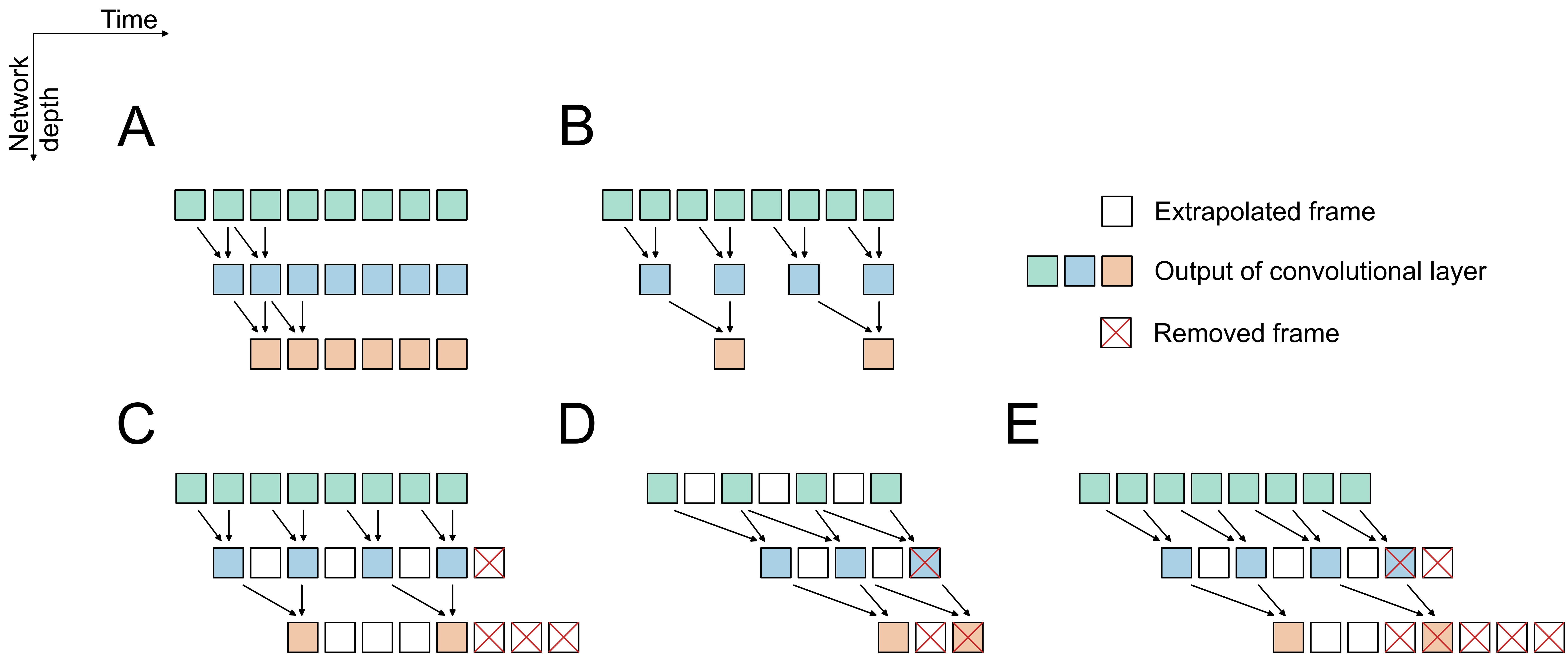}
    \else
        \includegraphics[width=0.85\textwidth]{images/methods/SOI_ops.png}
    \fi
    \caption{SOI for convolutional operations. For visualization purposes we show data as frames in time domain. A) Standard convolution. B) Strided convolution. C) Strided-Cloned Convolution. D) Shifted convolution. E) Shifted Strided-Cloned Convolution.}
    \label{fig:SOI_ops}
\end{figure*}

In this study, we introduce a novel method named Scattered Online Inference (SOI), which is based on the following key principles:

\begin{itemize}
    \item The reduction of computational complexity is achieved through the implementation of partial predictions of the network's future state.
    \item SOI operates as a two-phase system. The initial phase involves compressing data within the time domain using data compression. The subsequent phase focuses on data reconstruction, employing the most suitable extrapolation scheme.
    \item The method preserves the causal nature of the optimized network architecture.
    \item SOI's applicability is confined to a single-frame online inference and it necessitates the incorporation of skip connections to update the network's output following each inference.
\end{itemize}

\subsection{Limitations}

While our method, demonstrates significant advantages in reducing computational complexity, it is essential to acknowledge several limitations inherent to our approach.

First, the performance reduction observed with SOI, although acceptable within the context of our work, may not be tolerable in applications requiring the highest accuracy levels. This reduction, while compensated by a substantial reduction in computational cost, suggests a trade-off between efficiency and model performance that may limit SOI's applicability in scenarios where performance is the absolute priority.

Second, the flexibility provided by SOI to balance computational cost and model performance introduces complexity in selecting the optimal configuration. The method allows users to adjust this trade-off based on application requirements, but this demands careful tuning and validation, which could be resource-intensive in training. 

Third, the method's reliance on partial state predictions and data compression may introduce cumulative errors over time, particularly in longer sequences of time-series data. This is less critical in real-time, short-sequence applications but could degrade performance in tasks requiring continuous operation over extended periods without full model recalculations. 

Lastly, SOI's applicability is primarily demonstrated on specific neural network architectures. Although we expect SOI to generalize to other architectures, its effectiveness in reducing computational complexity while maintaining acceptable performance may vary depending on the network design, the nature of the task, and the dataset. Further research and experimentation are necessary to explore its utility across a broader range of models and applications.

\section{Methods}

When processing a time-series in online mode, the model goes through each incoming element\footnote{We define an ``element'' as a chunk of data of arbitrary size, including the smallest possible unit - a singular data point in a time series. The size of the incoming data chunk depends on the specific system being used.} separately. In this paper we refer to such an event as \emph{inference}.

An improvement in computation complexity is achieved by introducing the partially predictive compression layer adapted for online inference, as well as by avoiding the redundant computations done during previous inferences as in STMC study. Therefore, after each inference, the results which would be recalculated in subsequent runs are cached. We refer to such cacheable outputs as a \emph{partial state} of the network.

\subsection{Scattered Online Inference\label{section:SOI}}

Scattered Online Inference (SOI) is a method which modifies the inference pattern of a network to skip the recalculation of certain layers in a predetermined pattern. This is achieved through the use of compression and extrapolation. Both operations are exclusively applied along the time axis. In this study, as example, we employed strided convolutions as compression layers and a basic form of extrapolation, where the last known value is used to replace the missing ones\footnote{More sophisticated methods were tested as well and are presented in supplementary materials.}. To facilitate a better understanding of the SOI algorithm in CNNs, in Figure \ref{fig:SOI_ops}, we define three types of convolutional layers utilized in our method. For comparison purposes, Figure \ref{fig:SOI_ops} also includes standard convolution and strided convolution.

\begin{figure*}[t]
    \centering
    \ifdefined\enableColors
        \includegraphics[width=0.90\textwidth]{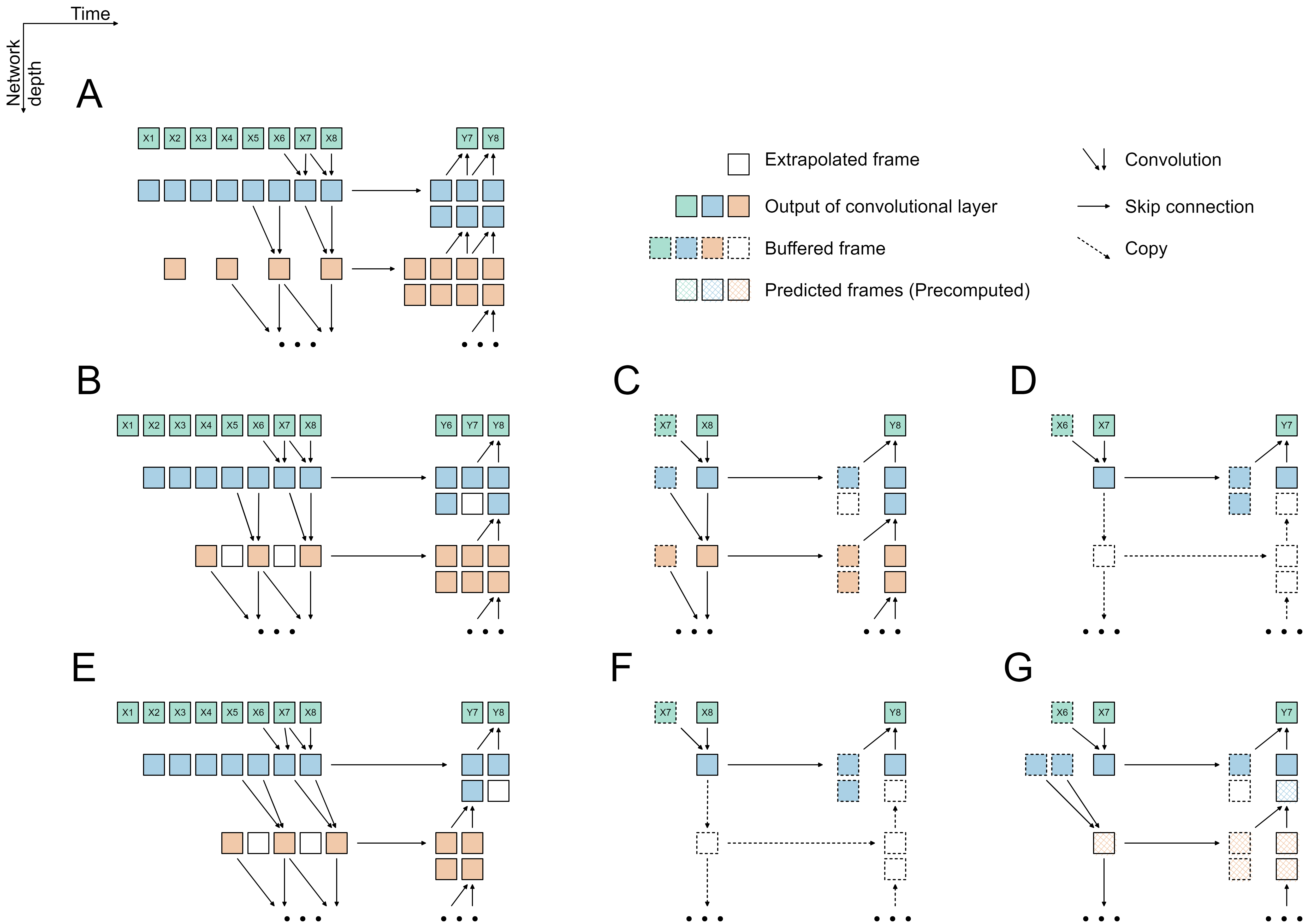}
    \else
        \includegraphics[width=0.90\textwidth]{images/methods/SOI_types.png}
    \fi
    \caption{Inference patterns of each type of SOI based on U-Net architecture. A) Unmodified causal U-Net. B) Partially predictive (PP) SOI. C) Even inference of PP. D) Odd inference of PP. E) Fully predictive (FP) SOI. F) Even inference of FP. G) Odd inference of FP.}
    \label{fig:SOI_types}
\end{figure*}

Strided-Cloned Convolution (S-CC) performs a 2-step operation. Firstly, it applies a strided convolution to the input in order to compress the data. In the second step it fills the gaps created by striding using any form of extrapolation. In our experiments we extrapolate by simply copying a previous frame. The copied frame is then aligned with a future frame relatively to its input. In practice, we split the stride and extrapolation operations into different layers which results in optimization of computational complexity between those layers. Because of that we will refer to this as S-CC pair.

Shifted Convolution (SC) shifts data in time domain after the convolution thus creating additional predicted network states. This layer may be used for additional latency reduction.

Shifted Strided-Cloned Convolution (SS-CC) is a combination of S-CC pair and SC layer which is needed if we want to do both of them at the same point of the network. In our experiments we extrapolate output vector first and then apply data shifting to reduce size of introduced partial state prediction.

SOI can be divided into two types depending on how prediction is handled. These two types have significantly different inference patterns (Fig. \ref{fig:SOI_types}).

\paragraph{Partially Predictive (PP)} In this type of SOI, we do not introduce any additional predictive components in the network. This implies that after compression, the most recent frame stores information for the current output frame and the future output frame. This type of SOI uses S-CC pairs only. This configuration results in only one inference (the first one), which updates all network states, while all subsequent ones use relevant buffered partial network states. Although hybrid setups involving more full passes are viable, they fall outside the scope of this paper. It's important to note that this mode does not reduce peak computational complexity, but rather the average computational complexity.

\paragraph{Fully Predictive (FP)} In this type of SOI, we introduce additional predictive components to the network. Compared to PP SOI, the most recent frame does not store any information about the current output frame, but rather about two future ones, hence the name ``Fully Predictive'', as only the future output is calculated. This is a more challenging task than PP SOI, but it can significantly decrease the latency of the model. This mode utilizes both S-CC pairs and SC layers. It optimizes both peak computational complexity and latency because it allows some inferences to be predicted in full. The fully predicted inference, in contrast to the regular inference which requires newly collected input, operates only on already processed data and can calculate relevant network states while the system awaits the new data, reducing the amount of computation required when the new data arrives.

Both of these types of SOI can be combined. This occurs when we first introduce PP SOI compression and then after some number of layers, we introduce an additional shift in the time axis after which the model can be treated as FP SOI.

\subsection{Mathematical Formulation\label{section:math}}

\newcommand{\concat}[3]{\left(#1 \;\middle|_{#2}\; #3 \right) }
\newcommand{\concatt}[2]{\concat{#1}{t}{#2} }
\newcommand{\concatc}[2]{\concat{#1}{c}{#2} }
\newcommand{\R}{\mathbb{R}}

Let us assume that an input of the model is represented by a 1D time series vector $X \in \mathbb{R}^N$ composed of $N$ samples. Additionally let's say that a network is composed of 5 convolutional layers and output vector $\prescript{l}{}{Y}$  for $l$-th layer is of the same shape as the input. Each convolutional layer has a 1D kernel $h_l \in \mathbb{R}^{M_l}$ which can be represented by a Toeplitz matrix $H_l \in \mathbb{R}^{N\times (N-M_l + 1)}$. We get:
\begin{equation}
\prescript{l}{}{Y} = H_l \cdot \prescript{l}{}{X}^T   
\end{equation}
After which we apply activation function $\sigma$ and get the input for the next layer:
\begin{equation}
\prescript{l+1}{}{X} = \sigma(\prescript{l}{}{Y})   
\end{equation}
We use $\prescript{l}{}{X}_t$ to denote a segment of $X$ ending in time $t$ of a length matching the context it is used in (e.g. assuming $h_l$ denotes a convolutional kernel of layer $l$ and provided with the context $h_l \cdot \prescript{l}{}{X}_t$, $\prescript{l}{}{X}_t$ has $M_l$ elements to match the kernel size).

By using the STMC inference pattern we perform inference for each element of $X$ separately and reuse past partial states of each layer to fully reconstruct the receptive field of the network which can be represented as follows:
\begin{equation}
\prescript{l+1}{}{X}_t = \concatt{\prescript{l+1}{}{X}_{t-1}}{\sigma(h_l \cdot {\prescript{l}{}{X}^T_t})}
\end{equation}
where $\concatt{\cdot}{\cdot}$ represents the concatenation of vectors in time axis.

If we use a convolutional layer with a stride of 2 as our second layer, then in the standard pattern, the size of the output vector of this layer is halved compared to its input. Consequently, every subsequent layer also has its input and output tensors reduced to half the size. We can restore the output to its original size by, for instance, applying transposed convolution. Let's assume that we apply transposed convolution in the 4th layer of our network. In comparison to our initial plain network, the new strided network will have the same number of operations in both the first and last layers. The 2nd, 3rd, and 4th layers will each have half the computations as before. In the 2nd layer, this reduction is due to the application of stride. In the third layer, it is a result of the smaller input size. Similarly, in the 4th layer, if we disregard multiplications by zero. 

When employing the STMC inference method, it's anticipated that each layer should process a single element and produce a single output. If we apply a similar stride and transposed convolution pattern without any additional modification, we'll face difficulties in reconstructing the output. Strided convolution would provide output values for even-numbered inferences (assuming a stride size of 2). However, during odd inferences, the 4th layer (transposed convolution) would require an upcoming even-numbered inference value, which would not yet be available. The authors of STMC propose a solution where every inference is treated as even-numbered, maintaining separate states for odd and even input positions. However, this pattern presents a challenge due to the exponential increase in the number of states for each added strided convolution.

SOI addresses this issue by removing the necessity of storing additional states, albeit at a cost to measured performance. For instance, in our network, we achieve this by abstaining from calculating the outputs of the 2nd, 3rd, and 4th layers during every even inference. To maintain causality, the transposed convolution output must be temporally shifted to produce either current and future frames or solely future frames, depending on the chosen SOI inference mode. Additionally, we advocate for the use of a skip connection between the input of the strided convolution and the output of the transposed convolution to update deeper layers of the network with information about the current data. This operation aims to minimize the influence of data forecasting on the optimized part of the network.

To formally describe partially predictive SOI, let's assume that the network contains layer $l_d$ with a stride of size 2\footnote{The choice of 2 is for notational simplicity. The same derivation can be applied for an arbitrary stride size.} and layer $l_u$ which reverses the downsampling performed by $l_d$. Typically, $l_d$ would be in the encoder part, while $l_u$ would be in the decoder.

The downsampling layer only returns a new element for even-numbered inferences. It's important to note that until the upsampling layer is reached, there is no need to perform any further computations when $t$ is odd. Namely for $l$ such that $l_d <l\leq l_u$:
\begin{wrapfigure}[24]{r}{4.6cm}
      \vspace{-0.25\baselineskip}
      \begin{center}
      \ifdefined\enableColors
          \includegraphics[width=0.36\textwidth]{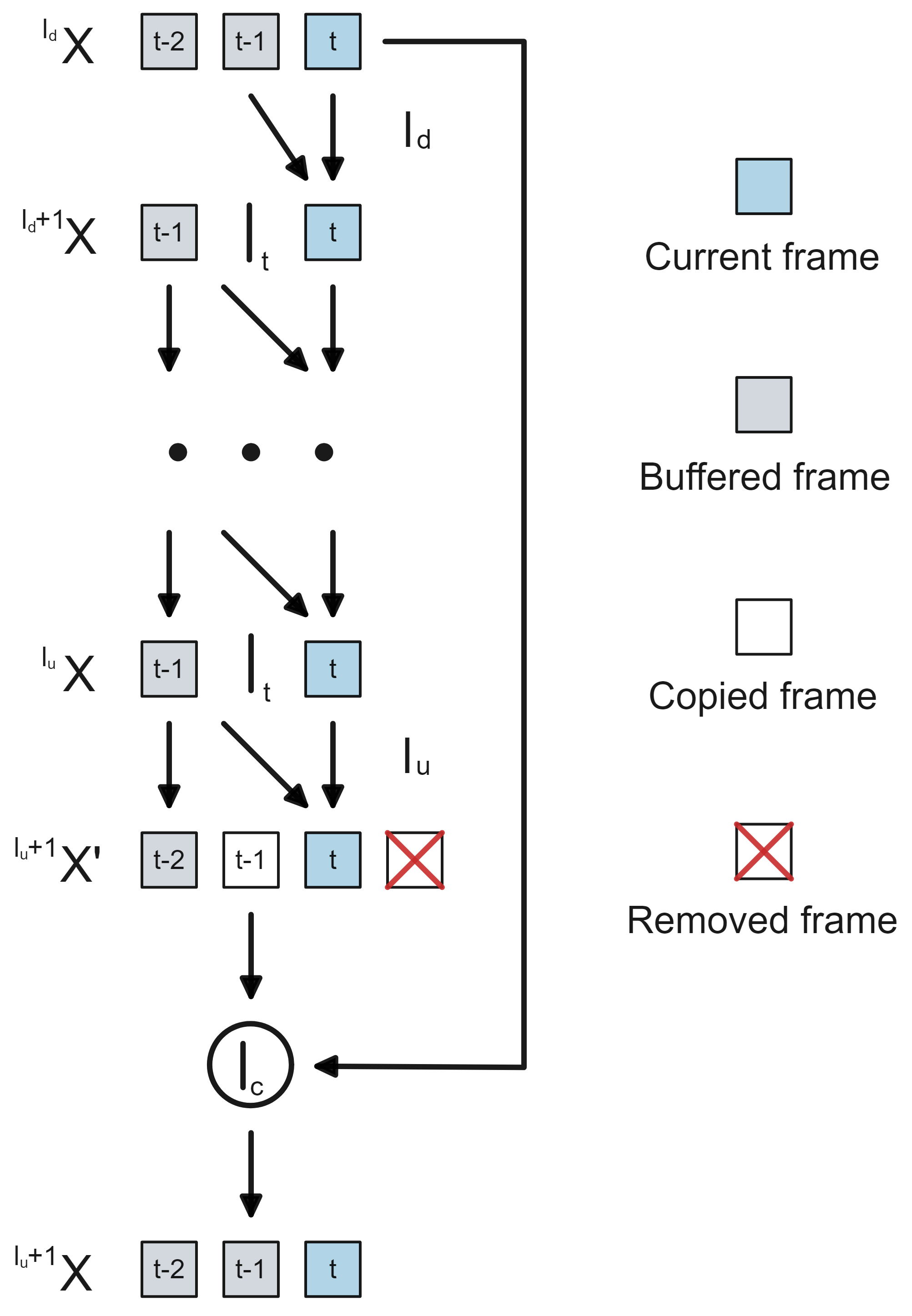}
      \else
          \includegraphics[width=0.36\textwidth]{images/methods/equations_flipped.png}
      \fi
      \end{center}
      \caption{SOI PP inference pattern.}
      \label{fig:equations}
\end{wrapfigure}\leavevmode
\begin{equation}
\prescript{l+1}{}{X}_{t} =  \begin{cases}
      \concatt{\prescript{l+1}{}{X}_{t-1}}{\sigma(h_{l} \cdot \prescript{l}{}{X}^T_{t})}, & \text{if}\ t\ \text{is even}\\
      \prescript{l+1}{}{X}_{t-1}, & \text{if}\ t\ \text{is odd}
    \end{cases}
\end{equation}

At layer $l_u$ we reconstruct the output by duplicating the convolution output.
\begin{equation}
\prescript{l_u+1}{}{X}_{t}' = \concatt{\prescript{l_u+1}{}{X}_{t-1}'}{\sigma(h_{l_u} \cdot \prescript{l_u}{}{X}^T_{t} )^T}
\label{eq:extrapolation}
\end{equation}
Note that $\prescript{l_u}{}{X}^T_{2s} = \prescript{l_u}{}{X}^T_{2s+1}$.

We then concatenate the output with data from the skip connection ($\concatc{\cdot}{\cdot}$ represents the concatenation of vectors in channel axis).
\begin{equation}
\prescript{l_u+1}{}{X}_{t} = \concatc{\prescript{l_u+1}{}{X}_{t}'}{\prescript{l_d}{}{X}_{t}}
\end{equation}
In the example above we traded inference over $l_u - l_d$ layers at the cost of inference over additional channels from skip connection concatenation. The additional cost of skip connection does not take effect in architectures where skip connections naturally exist like U-Net which we use as one of the examples in this study.

For fully predictive variant we modify equation (\ref{eq:extrapolation}) and add a shift in time axis. 
\begin{equation}
\prescript{l_u+1}{}{X}_{t}' = \concatt{\prescript{l_u+1}{}{X}_{t-1}'}{\sigma(h_{l_u} \cdot \prescript{l_u}{}{X}^T_{t-1} )^T}
\end{equation}
SOI can be seen as a form of forecasting where instead of training the model to predict output for input yet unknown, we predict partial states of the network. This is because preserving causality in the strided convolution forces us to predict a convolution results when the next second input is not yet available.

\section{Experiments}

\subsection{Speech Separation\label{section:SE}}

We selected speech separation as our first experimental task. The choice of the task is dictated by our current research interests and the potential benefits of fast online inference. In this task we focus on separating clean speech signal from noisy background. In literature this task is also referred to as speech denoising or noise suppression.

For this experiment we adopted the U-Net architecture as it is widely used for this specific task and inherently has skip connections which will allow for applying SOI inference pattern without substantial alterations. Our model is composed of 7 encoder and 7 decoder layers. The Deep Noise Suppression (DNS) Challenge - Interspeech 2020 dataset \citep{reddy2020}, licensed under CC-BY 4.0, was used for both training and evaluation purposes.

\paragraph{Position of S-CC pair} 
By introducing S-CC pair to the network we are enforcing data predictiveness of the network. The exact number of the predicted future partial states of the model depends on the position of S-CC pair and number of those pairs within the network. In addition it is worth noting that larger amount of predicted partial states of the network leads to higher reduction of computational complexity. In this experiment we test every position of S-CC pairs while applying up to two such modules to the model's architecture.

\paragraph{Position of SS-CC pair}
SS-CC pair introduces additional shift in time axis compered to S-CC pair. In this experiment we alter the position of SS-CC pair within the network and also separately alter the position of S-CC pair and time shift which might be consider as a hybrid of partially and fully predictive pattern.

\paragraph{Resampling}
Simple resampling of audio signal may be used to reduce the number of calculations done by neural networks but will yield significant increase in model latency. Nonetheless in this experiment we compare SOI to four different resampling methods: SoX which is based on method proposed by \cite{Laurent2004}, using Kaiser window, polyphase and linear. With this methods we resampled our speech separation dataset from 16k to 8k at the input of the model and then from 8k to 16k at the output of the model. For every resampling we used our baseline model and compared it to three selected SOI models.

\paragraph{Pruning}
In this experiment we used unstructured global magnitude pruning (similar to \cite{Han2015_learning}) to show how our method can be combined with pruning leading to better results than pruning on a standard model. For this experiment we chose to prune ``SOI 1'' and ``SOI 2$\vert$6'' variants of our baseline model. Each step we pruned 4096 weights from model and we report how models performed on each step.

\subsection{Acoustic Scene Classification}
Acoustic scene classification (ASC) is our second task of choice. The goal of the task is to estimate the location where the specific sample was recorded. This task is commonly considered as an auxiliary problem in various online scenarios such as  selection of bank of denoising filters. 

For all our tests in ASC task we used GhostNet architecture \citep{Han_2020_CVPR}. We tested 7 different model sizes for all 3 variants - Baseline, STMC and SOI. Each test was repeated 5 times. We used the TAU Urban Acoustic Scene 2020 Mobile dataset \citep{heittola_toni_2020_3819968} for both training and validation. Models were trained on a single Nvidia P40 GPU for 500 epochs using Adam optimizer with initial learning rate of 1e-3. 

\section{Results}

\subsection{Speech Separation}

\paragraph{Partially Predictive}
Results of partially predictive SOI in speech separation task are shown in figure \ref{fig:SOI_PP_results}. We tested variants with a single S-CC layer (``S-CC'') and two S-CC layers (``2xS-CC $X$''). In the latter case, the experiments are grouped by the position of the first S-CC layer ($X$). Value of SI-SNRi metric is highly dependant on the position of S-CC layers. Generally, the earlier the S-CC layer is introduced the lower SI-SNRi but higher complexity reduction. This phenomena may be explained by the difficulty of the partial state prediction task. All layers subsequent to the S-CC layer are required to predict twice the number of elements compared to the layers without having the S-CC as an prior layer. Overall, with SOI we can achieve linear dependency between computation cost and SI-SNRi up to 64\% of complexity reduction (with linearity factor of 0.001 dB SI-SNRi per 1 MMAC/s (0.017 dB SI-SNRi per 1\% of reduction)). A significant SI-SNRi drop is observed if S-CC layer is introduced too early. In table \ref{table:results_PP} we present selected results from the whole experiment.

\paragraph{Fully Predictive}
Results for speech separation using fully predictive SOI are presented in figure \ref{fig:STC_vs_SOI}. Details for selected models can be found in table \ref{table:results_FP}. Reduction of metrics observed for fully predictive variant tends to be larger than for a single S-CC layer in partially predictive SOI. Additional reduction of metrics stems from added shift in time axis. Added shift produces a pattern where some part of model can be updated between inferences, as it only depends on past data. Size of this model part depends on the position of an SC or SS-CC layer within network. We refer to the size of this model part as ``Precomputed'' in table \ref{table:results_FP}. The increase of precomputed partial state leads to reduction of SI-SNRi metric but allows to speed-up computation, because this part can be computed in advance. With FP SOI we achieved 50\% reduction of computational cost at the expense of 11.3\% of SI-SNRi but with 83.7\% of network calculated using past data only.

\begin{table}[H]
  \begin{minipage}{.45\linewidth}
      \centering
      \ifdefined\enableColors
          \includegraphics[width=1\textwidth]{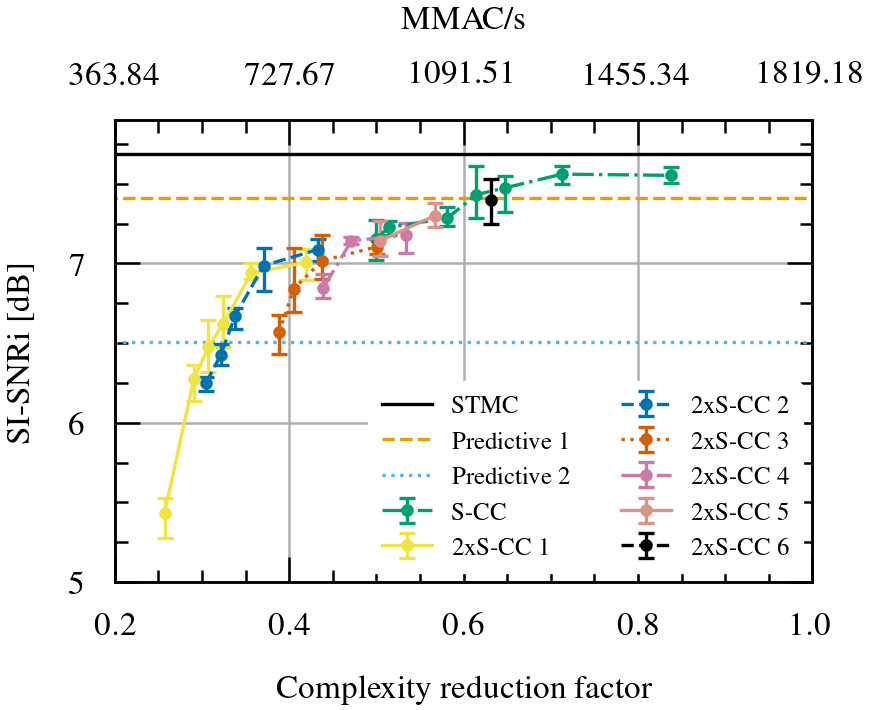}
      \else
          \includegraphics[width=1\textwidth]{images/evaluation/combined_si_snri_vs_mean_complexity_reduction.png}
      \fi
      \captionof{figure}
        {%
          Results of speech separation experiment with PP SOI.%
          \label{fig:SOI_PP_results}%
        }
  \end{minipage}\hspace{10mm}
  \begin{minipage}{.46\linewidth}
  \vspace{-6mm}
  \captionof{table}
        {%
          Selected results from experiments with partially predictive SOI for speech separation.%
          \label{table:results_PP}%
        }
    \centering
    \tiny
    {
    \setlength{\tabcolsep}{3pt}
    \begin{tabular}{l c S[table-format=3.1] S[table-format=3.1] S[table-format=4.1]}
      \toprule
      Model               & \multicolumn{1}{c}{SI-SNRi}   & \multicolumn{1}{c}{SI-SNRi}             & \multicolumn{1}{c}{Complexity}         & \multicolumn{1}{c}{Complexity} \\
                          & \multicolumn{1}{c}{(dB)}      & \multicolumn{1}{c}{retain (\%)}    & \multicolumn{1}{c}{retain (\%)}   & \multicolumn{1}{c}{(MMAC/s)}     \\
      \midrule
      STMC                & \textbf{7.69\SPSB{\,\fontsize{3}{1}\selectfont +0.06}{\,\fontsize{3}{1}\selectfont -0.08}} & {\bfseries 100.0} & 100.0         & 1819.2                          \\
      Predictive 1        & 7.41\SPSB{\,\fontsize{3}{1}\selectfont +0.07}{\,\fontsize{3}{1}\selectfont -0.09} &  96.3 & 100.0                          & 1819.2                          \\
      Predictive 2        & 6.51\SPSB{\,\fontsize{3}{1}\selectfont +0.03}{\,\fontsize{3}{1}\selectfont -0.08} &  84.7 & 100.0                          & 1819.2                          \\
      \midrule                      
      S-CC 2              & 7.23\SPSB{\,\fontsize{3}{1}\selectfont +0.04}{\,\fontsize{3}{1}\selectfont -0.05} &  94.0  & 51.4                           & 935.2                          \\
      S-CC 5              & 7.47\SPSB{\,\fontsize{3}{1}\selectfont +0.07}{\,\fontsize{3}{1}\selectfont -0.15} &  97.2  & 64.8                           & 1178.7                          \\
      S-CC 7              & 7.55\SPSB{\,\fontsize{3}{1}\selectfont +0.05}{\,\fontsize{3}{1}\selectfont -0.05} &  98.2& 83.8                           & 1524.3   \\
      \midrule
      2xS-CC 1$\vert$3    & 6.27\SPSB{\,\fontsize{3}{1}\selectfont +0.09}{\,\fontsize{3}{1}\selectfont -0.14} &  81.6 & 29.1                  & 528.8                           \\
      2xS-CC 1$\vert$6    & 6.94\SPSB{\,\fontsize{3}{1}\selectfont +0.07}{\,\fontsize{3}{1}\selectfont -0.03} &  90.2 & 35.6                           & 648.5                           \\
      2xS-CC 2$\vert$5    & 6.67\SPSB{\,\fontsize{3}{1}\selectfont +0.05}{\,\fontsize{3}{1}\selectfont -0.09} &  86.8 & 33.8                           & 615.0                           \\
      2xS-CC 3$\vert$6    & 7.02\SPSB{\,\fontsize{3}{1}\selectfont +0.16}{\,\fontsize{3}{1}\selectfont -0.12} &  91.3 & 43.8                           & 796.4                           \\
      2xS-CC 4$\vert$6    & 7.14\SPSB{\,\fontsize{3}{1}\selectfont +0.02}{\,\fontsize{3}{1}\selectfont -0.02} &  92.9 & 47.1                           & 857.3                           \\
      2xS-CC 5$\vert$7    & 7.30\SPSB{\,\fontsize{3}{1}\selectfont +0.08}{\,\fontsize{3}{1}\selectfont -0.07} &  94.9 & 56.7                           & 1031.2                            \\
      2xS-CC 6$\vert$7    & 7.40\SPSB{\,\fontsize{3}{1}\selectfont +0.13}{\,\fontsize{3}{1}\selectfont -0.15} &  96.2 & 63.2                           & 1149.5                            \\
      \bottomrule
    \end{tabular}
    }
    \end{minipage}
  \end{table}

\begin{table}[H]
\vspace{-7.5mm}
  \begin{minipage}{.4\linewidth}
      \centering
      \ifdefined\enableColors
          \includegraphics[width=1\textwidth]{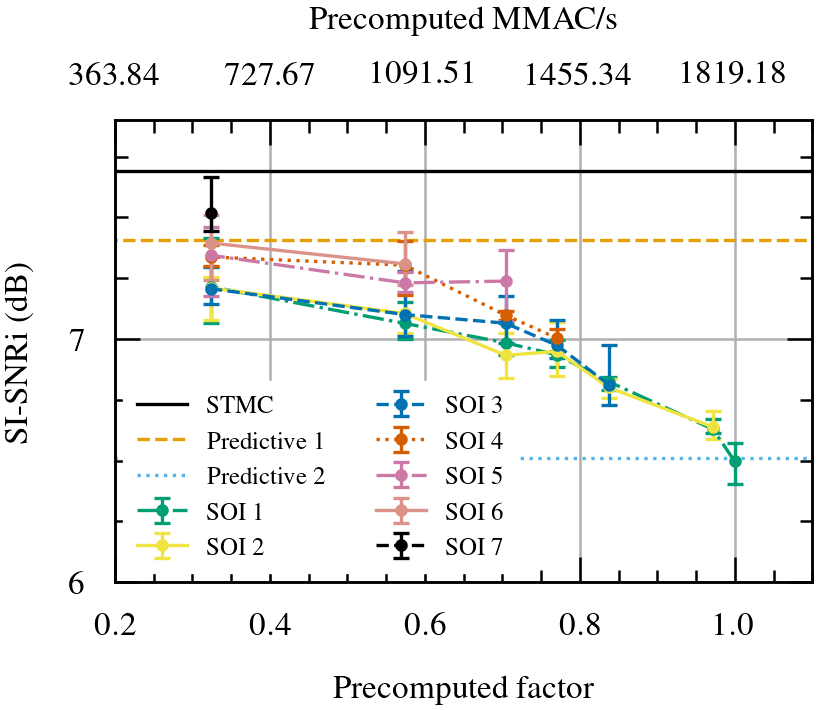}
      \else
          \includegraphics[width=1\textwidth]{images/evaluation/FP_combined_mean_si_snri_vs_complexity_reduction.png}
      \fi
          \captionof{figure}
        {%
          Results of speech separation experiment with FP SOI.%
          \label{fig:STC_vs_SOI}%
        }
  \end{minipage}\hspace{10mm}
  \begin{minipage}{.5\linewidth}
  \vspace{-5mm}
  
    \captionof{table}
        {%
          Selected results from experiments with fully predictive SOI for speech separation.%
          \label{table:results_FP}%
        }
    \centering
    \tiny
    {
    \setlength{\tabcolsep}{3pt}
    \begin{tabular}{l c S[table-format=3.1] S[table-format=3.1] S[table-format=4.1] S[table-format=3.1]}
      \toprule
      Model               & \multicolumn{1}{c}{SI-SNRi}   & \multicolumn{1}{c}{SI-SNRi}   & \multicolumn{1}{c}{Complexity}        & \multicolumn{1}{c}{Complexity} & \multicolumn{1}{c}{Precomputed}\\
                          & \multicolumn{1}{c}{(dB)}      & \multicolumn{1}{c}{retain (\%)}   &  \multicolumn{1}{c}{retain (\%)}  & \multicolumn{1}{c}{(MMAC/s)} & \multicolumn{1}{c}{(\%)}      \\
      \midrule
      STMC                & \textbf{7.69\SPSB{\,\fontsize{3}{1}\selectfont +0.06}{\,\fontsize{3}{1}\selectfont -0.08}} & {\bfseries 100.0} & 100.0        & 1819.2 & 0.0                          \\
      Predictive 1        & 7.41\SPSB{\,\fontsize{3}{1}\selectfont +0.07}{\,\fontsize{3}{1}\selectfont -0.09} &  96.3 & 100.0                          & 1819.2 & {\bfseries100.0}                          \\
      Predictive 2        & 6.51\SPSB{\,\fontsize{3}{1}\selectfont +0.03}{\,\fontsize{3}{1}\selectfont -0.08} &  84.7 & 100.0                         & 1819.2 & {\bfseries100.0}                     \\
      \midrule
      SS-CC 2           & 6.64\SPSB{\,\fontsize{3}{1}\selectfont +0.07}{\,\fontsize{3}{1}\selectfont -0.05} &  86.3 & 51.4                          & 935.2   & 97.2                   \\
      SS-CC 5           & 7.24\SPSB{\,\fontsize{3}{1}\selectfont +0.13}{\,\fontsize{3}{1}\selectfont -0.16} &  94.1 & 64.8                           & 1178.7       & 70.4                    \\
      SS-CC 7    & 7.52\SPSB{\,\fontsize{3}{1}\selectfont +0.15}{\,\fontsize{3}{1}\selectfont -0.07} &  97.8 & 83.8                           & 1524.3       & 32.4                       \\
      \midrule
      S-CC 1$\vert$3    & 6.82\SPSB{\,\fontsize{3}{1}\selectfont +0.02}{\,\fontsize{3}{1}\selectfont -0.04} &  88.7 & {\bfseries50.0}                          &  {\bfseries 909.6} & 83.7                       \\
      S-CC 1$\vert$6    & 7.06\SPSB{\,\fontsize{3}{1}\selectfont +0.09}{\,\fontsize{3}{1}\selectfont -0.06} &  91.8 & {\bfseries50.0}                         & {\bfseries 909.6}    & 57.4                       \\
      S-CC 2$\vert$5    & 6.93\SPSB{\,\fontsize{3}{1}\selectfont +0.09}{\,\fontsize{3}{1}\selectfont -0.09} &  90.1 & 51.4                           & 935.2     & 70.4                        \\
      S-CC 3$\vert$6    & 7.10\SPSB{\,\fontsize{3}{1}\selectfont +0.18}{\,\fontsize{3}{1}\selectfont -0.09} &  92.3 & 58.1                           & 1057.5     & 57.4                         \\
      S-CC 4$\vert$6    & 7.30\SPSB{\,\fontsize{3}{1}\selectfont +0.10}{\,\fontsize{3}{1}\selectfont -0.12} &  94.9 & 61.5                          & 1118.4      & 57.4                       \\
      S-CC 5$\vert$6    & 7.23\SPSB{\,\fontsize{3}{1}\selectfont +0.05}{\,\fontsize{3}{1}\selectfont -0.04} &  94.0 & 64.8                           & 1178.7         & 57.4                  \\
      S-CC 6$\vert$7    & 7.39\SPSB{\,\fontsize{3}{1}\selectfont +0.12}{\,\fontsize{3}{1}\selectfont -0.15} &  96.1 & 71.3                           & 1296.9        & 32.4                     \\
      \bottomrule
    \end{tabular}
    }
  \end{minipage}
\end{table}

\paragraph{Resampling}
The results presented in the table \ref{table:soi_resampling} show that SOI outperforms the listed resampling methods in terms of complexity reduction under the constrained of preserving the quality of the original model (STMC). The results for resampling-based complexity reduction indicate that model’s quality depends on the used resampling algorithm and, for all the selected ones, the initial information loss strongly affects the model performance.

\begin{table}[H]
  \begin{minipage}{.5\linewidth}
      \centering
      \ifdefined\enableColors
          \includegraphics[width=0.8\textwidth]{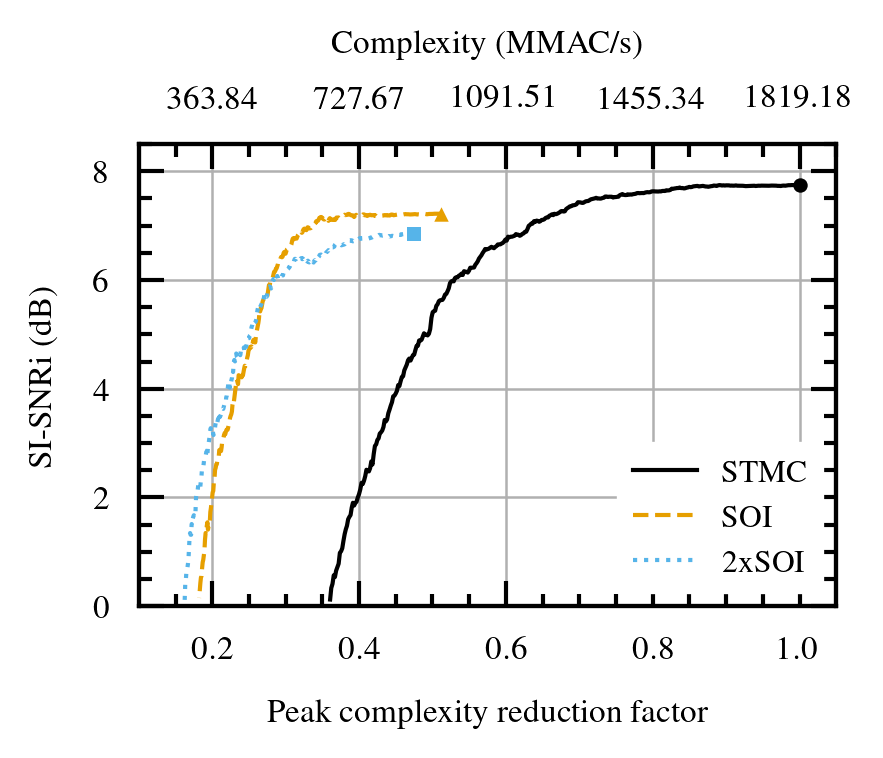}
      \else
          \includegraphics[width=0.8\textwidth]{images/appendix/pruning.png}
      \fi
          \captionof{figure}
        {%
          Pruning of STMC, SOI and 2xSOI models. Unpruned models are indicated by markers.%
          \label{fig:pruning}%
        }
  \end{minipage}\hspace{4mm}
  \begin{minipage}{.45\linewidth}
  \vspace{-10mm}
  
    \captionof{table}
        {%
          Comparison between resampling and SOI.%
          \label{table:soi_resampling}%
        }
    \centering
    \tiny
    {
    \setlength{\tabcolsep}{3pt}
    \begin{tabular}{l@{\hskip 5mm} c@{\hskip 5mm} c}
      \toprule
      \multirow{2}{*}{Method} & SI-SNRi & Complexity  \\
                              & (dB)    & (MMAC/s) \\
      \midrule
      STMC & 7.69 & 1819.2 \\
      \midrule
      Linear & 3.49 & 909.6 \\[1mm]
      Polyphase & 5.69 & 909.6 \\[1mm]
      Kaiser & 5.83 & 909.6 \\[1mm]
      SoX & 5.77 &  909.6 \\
      \midrule
      S-CC 5 & \textbf{7.47} & 1178.7 \\[1mm]
      S-CC 2 & 7.23 & 935.2 \\[1mm]
      S-CC 1$|$3 & 6.27 & \textbf{528.8} \\
      \bottomrule
    \end{tabular}
    }
  \end{minipage}
\end{table}

\paragraph{Pruning}
The application of SOI along with the pruning of STMC model surpasses the effect of application of the pruning alone. The addition of SOI allowed for achieving a further reduction in computational complexity by around 300 MMAC/s for the same model’s performance, which is about 16\% of the original model. Interestingly the “SOI 2|6” surpassed the “SOI 1” model at around 6 dB SI-SNRi. The results of the experiment are shown in Fig. \ref{fig:pruning}.

\subsection{Acoustic Scene Classification}

\begin{wraptable}[19]{r}{7.0cm}
    \vspace{-0.8cm}
\caption{Results of ASC experiment.}
\centering
{
\setlength{\tabcolsep}{3pt}
\tiny
\begin{tabular}{c c c c c}
\toprule
\multirow{2}{*}{Model} & \multirow{2}{*}{Method} & Top-1         & Complexity & Number of  \\
                       &                         & Accuracy (\%) & (MMAC/s)   & Parameters \\
\midrule
\multirow{3}{*}{I}   & Baseline &         55.68\SPSB{\,\fontsize{3}{1}\selectfont +0.62}{\,\fontsize{3}{1}\selectfont -0.96}  &       423.07   &  \textbf{1470} \\
                     & STMC     &         55.68\SPSB{\,\fontsize{3}{1}\selectfont +0.62}{\,\fontsize{3}{1}\selectfont -0.96}  &         0.41   &  \textbf{1470} \\
                     & SOI      & \textbf{55.90\SPSB{\,\fontsize{3}{1}\selectfont +1.83}{\,\fontsize{3}{1}\selectfont -5.61}} & \textbf{0.37}  &          1833  \\
\midrule
\multirow{3}{*}{II}  & Baseline & \textbf{64.18\SPSB{\,\fontsize{3}{1}\selectfont +0.72}{\,\fontsize{3}{1}\selectfont -0.43}}  &       959.67   &  \textbf{3352} \\
                     & STMC     & \textbf{64.18\SPSB{\,\fontsize{3}{1}\selectfont +0.72}{\,\fontsize{3}{1}\selectfont -0.43}}  &         0.94   &  \textbf{3352} \\
                     & SOI      &         61.98\SPSB{\,\fontsize{3}{1}\selectfont +1.06}{\,\fontsize{3}{1}\selectfont -1.38}   & \textbf{0.80}  &          3594  \\
\midrule
\multirow{3}{*}{III} & Baseline &         66.45\SPSB{\,\fontsize{3}{1}\selectfont +1.75}{\,\fontsize{3}{1}\selectfont -1.69}  &      1624.11   &  \textbf{5814} \\
                     & STMC     &         66.45\SPSB{\,\fontsize{3}{1}\selectfont +1.75}{\,\fontsize{3}{1}\selectfont -1.69}  &         1.59   &  \textbf{5814} \\
                     & SOI      & \textbf{68.14\SPSB{\,\fontsize{3}{1}\selectfont +0.92}{\,\fontsize{3}{1}\selectfont -2.23}} & \textbf{1.37}  &          6653  \\
\midrule
\multirow{3}{*}{IV}  & Baseline & \textbf{70.57\SPSB{\,\fontsize{3}{1}\selectfont +2.64}{\,\fontsize{3}{1}\selectfont -1.81}}  &      2405.09   &  \textbf{8696} \\
                     & STMC     & \textbf{70.57\SPSB{\,\fontsize{3}{1}\selectfont +2.64}{\,\fontsize{3}{1}\selectfont -1.81}}  &         2.35   &  \textbf{8696} \\
                     & SOI      &         70.32\SPSB{\,\fontsize{3}{1}\selectfont +1.17}{\,\fontsize{3}{1}\selectfont -1.98}   & \textbf{1.97}  &          8835  \\
\midrule
\multirow{3}{*}{V}   & Baseline & \textbf{76.91\SPSB{\,\fontsize{3}{1}\selectfont +0.74}{\,\fontsize{3}{1}\selectfont -1.40}} &      6769.78   &         25480  \\
                     & STMC     & \textbf{76.91\SPSB{\,\fontsize{3}{1}\selectfont +0.74}{\,\fontsize{3}{1}\selectfont -1.40}} &         6.61   &         25480  \\
                     & SOI      &         76.42\SPSB{\,\fontsize{3}{1}\selectfont +0.66}{\,\fontsize{3}{1}\selectfont -0.63}  & \textbf{5.54}  & \textbf{24632} \\
\midrule
\multirow{3}{*}{VI}  & Baseline & \textbf{81.66\SPSB{\,\fontsize{3}{1}\selectfont +1.00}{\,\fontsize{3}{1}\selectfont -0.98}}  &      13187.40  &         50392 \\
                     & STMC     & \textbf{81.66\SPSB{\,\fontsize{3}{1}\selectfont +1.00}{\,\fontsize{3}{1}\selectfont -0.98}}  &         12.78  &         50392 \\
                     & SOI      &         80.73\SPSB{\,\fontsize{3}{1}\selectfont +1.79}{\,\fontsize{3}{1}\selectfont -1.36}   & \textbf{10.75} & \textbf{47605}\\
                     \midrule
\multirow{3}{*}{VII} & Baseline &         83.07\SPSB{\,\fontsize{3}{1}\selectfont +1.03}{\,\fontsize{3}{1}\selectfont -0.68}  &      21395.26  & \       83432 \\
                     & STMC     &         83.07\SPSB{\,\fontsize{3}{1}\selectfont +1.03}{\,\fontsize{3}{1}\selectfont -0.68}  &         20.87  &         83432  \\
                     & SOI      & \textbf{83.35\SPSB{\,\fontsize{3}{1}\selectfont +0.74}{\,\fontsize{3}{1}\selectfont -1.12}} & \textbf{17.59} & \textbf{77753} \\                     
\bottomrule
\end{tabular}
}
\label{table:ASC_results}
\end{wraptable}

Results for the ASC task are collected in Table \ref{table:ASC_results}. For models I, III and VII we observed that our method led to an increase in accuracy, whereas other models showed a the decrease in metrics. The largest decrease in accuracy was around 2.20\%, and the largest increase was 1.69\%. These results indicate that SOI does not lead to a decrease in model quality for this particular task compared to the STMC model. This can be explained by the much slower change of output (acoustic scene label) compared to the previous task (speech mask). In our tests, the reduction in computational complexity of SOI models amounted to around 16\% compared to the STMC model. This reduction decreased to 11\% for the smallest model due to the addition of the skip connections. For this particular architecture, our method also reduced the number of parameters for the largest tested models by around 7\%.

\section{Conclusion}

In this work, we presented a method for reducing the computational cost of a convolutional neural network by reusing network partial states from previous inferences, leading to a generalization of these states over longer time periods. We discussed the effect of partial state prediction that our method imposes on the neural model and demonstrated that it can be applied gradually to balance model quality metrics and computational cost.

Our experiments highlight the high potential for computational cost reduction of a CNN, especially for tasks where the output remains relatively constant, such as event detection or classification. We achieved a computational cost reduction of 50\% without any drop in metrics in the ASC task and a 64.4\% reduction in computational cost with a relatively small reduction of 9.8\% in metrics for the speech separation task. We also showcased the ability of SOI to control the trade-off between model's quality and computational cost, allowing for resource- and requirement-aware tuning. 

The presented method offers an alternative to the STMC solution for strided convolution. While SOI reduces network computational complexity at the expense of measured performance, STMC ensures that metrics are not reduced but at the cost of increased memory consumption at an exponential rate. SOI is akin to methods like network pruning, but unlike pruning, it does not require special sparse kernels for inference optimization. It is worth noting that these methods are not mutually exclusive, therefore, the STMC strided convolution handler, SOI, and pruning can coexist within a neural network to achieve the desired model performance.

\bibliography{neurips_2024}
\bibliographystyle{iclr2024_conference}

\newpage
\appendix
\onecolumn
\relax

\section{Reproducibility Notes}

\subsection{Speech Separation}
For this experiment we adopted the U-Net architecture as it is widely used for this specific task and inherently has skip connections which will allow for applying SOI inference pattern without substantial alterations. Our model is composed of 7 encoder and 7 decoder layers, each comprising STMC/tSTMC, batch norm and ELU activation layers. Each model was trained for 100 epochs using Adam optimizer with initial learning rate of 1e-3. We trained each model on a single Nvidia P40 GPU 5 times and reported the average SI-SNRi. The mean training time of a single model amounted to about 14 hours. The Deep Noise Suppression (DNS) Challenge - Interspeech 2020 dataset \citep{reddy2020}, licensed under CC-BY 4.0, was used for both training and evaluation purposes. For training, we used 16384 10s samples without any form of augmentation and for both validation and test sets we used 64 samples with similar setup to the training set.

\subsection{Acoustic Scene Classification}
For all our tests in ASC task we used GhostNet architecture \citep{Han_2020_CVPR}. Our baseline model is the original architecture with “same” padding making it not applicable in online scenario. STMC model changes the padding to manually-applied padding in left-most (oldest) side of data and applies STMC inference pattern. SOI model adds upsampling after each processing block and skip connections between downsampling/upsampling layers.

Models were trained on a single Nvidia P40 GPU for 500 epochs using Adam optimizer with initial learning rate of 1e-3. We tested 7 different model sizes for all 3 variants - Baseline, STMC and SOI. Each test was repeated 5 times. The mean training time of a single model amounted to about 4 hours. We used the TAU Urban Acoustic Scene 2020 Mobile dataset \citep{heittola_toni_2020_3819968} for both training and validation.

\section{Strided Convolutions are Better for Longer Predictions \label{section:strided_conv_vs_prediction}}

\begin{wrapfigure}[14]{r}{6cm}
    \vspace{-0.5cm}
    \centering
    \includegraphics[width=5.8cm]{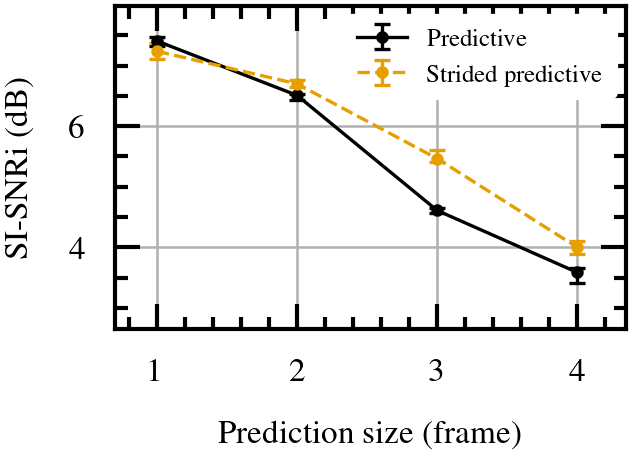}
    \caption{Comparison between standard convolutions and strided convolution in predictive inference.}
    \label{fig:predictive_inference}

\end{wrapfigure}

In this experiment, we investigated the impact of strided convolutions on predictive inference. Our test environment consisted of a U-Net model applied to a speech separation task. We examined two model variants: \emph{Predictive} and \emph{Strided Predictive}. The Predictive model serves as our baseline U-Net with an added time shift at the end. The Strided Predictive model, in addition to the time shift, incorporates strided convolutions in place of standard ones. For each model, we conducted tests with four different lengths of prediction, ranging from 1 frame to 4 frames. We performed five training runs for each model using the setup outlined in Section \ref{section:SE}. The results of this experiment are displayed in Figure \ref{fig:predictive_inference} and Table \ref{table:results_app_a}.\\

\vspace{-0.2cm}
We conclude that strided convolutions shows higher potential for longer predictions. We attribute this effect to the fact, that using strides forces stronger generalization of outputs of strided convolutions because they are applied in multiple contexts.
    
\begin{table}[h]
  \vspace{-0.4cm}
  \caption{Results of experiment on the influence of strided convolution on predictive inference.}
  \centering

  \setlength{\tabcolsep}{3pt}
  \begin{tabular}{c@{\hskip 5mm} c@{\hskip 5mm} c}
    \toprule
    Length of  & \multicolumn{2}{c}{SI-SNRi (dB)}     \\
    \cmidrule(lr){2-3}
    prediction & Predictive & Strided predictive \\
    \midrule
    1                    &  \textbf{7.41}\SPSB{\,\fontsize{6}{1}\selectfont +0.07}{\,\fontsize{6}{1}\selectfont -0.09}                                             &  7.24\SPSB{\,\fontsize{6}{1}\selectfont +0.14}{\,\fontsize{6}{1}\selectfont -0.13}       
                         \\[1mm]
    2                    &  6.51\SPSB{\,\fontsize{6}{1}\selectfont +0.03}{\,\fontsize{6}{1}\selectfont -0.08}                                             &  \textbf{6.70}\SPSB{\,\fontsize{6}{1}\selectfont +0.07}{\,\fontsize{6}{1}\selectfont -0.05}       
                         \\[1mm]
    3                    &  4.61\SPSB{\,\fontsize{6}{1}\selectfont +0.04}{\,\fontsize{6}{1}\selectfont -0.05}                                             &  \textbf{5.47}\SPSB{\,\fontsize{6}{1}\selectfont +0.14}{\,\fontsize{6}{1}\selectfont -0.07}       
                         \\[1mm]
    4                    &  3.59\SPSB{\,\fontsize{6}{1}\selectfont +0.08}{\,\fontsize{6}{1}\selectfont -0.17}                                             &  \textbf{4.00}\SPSB{\,\fontsize{6}{1}\selectfont +0.11}{\,\fontsize{6}{1}\selectfont -0.11}       
                         \\
    \bottomrule
  \end{tabular}
  \label{table:results_app_a}
\end{table}

\section{Inference Time and Peak Memory Footprint}

To showcase SOI's influence on inference time and peak memory consumption, we extended our results from Table \ref{table:results_PP} to include these measurements with a single S-CC layer. The collected results are presented in Table \ref{table:inferene_time_and_memory}.

\begin{table}[H]
  \caption{
      Results from experiments with partially predictive SOI for speech separation with added average inference time and peak memory footprint.%
    }
    \centering
    {
    \setlength{\tabcolsep}{4pt}
    \begin{tabular}{l c S[table-format=3.1] S[table-format=3.1] S[table-format=4.1] c S[table-format=2.1]}
      \toprule
      Model               & \multicolumn{1}{c}{SI-SNRi}   & \multicolumn{1}{c}{SI-SNRi}             & \multicolumn{1}{c}{Complexity}         & \multicolumn{1}{c}{Complexity} & \multicolumn{1}{c}{Avg. inference} & \multicolumn{1}{c}{Peak memory}  \\
                          & \multicolumn{1}{c}{(dB)}      & \multicolumn{1}{c}{retain (\%)}    & \multicolumn{1}{c}{retain (\%)}             & \multicolumn{1}{c}{(MMAC/s)}   & \multicolumn{1}{c}{time (ms)}   & \multicolumn{1}{c}{footprint (MB)}    \\
      \midrule
      STMC                & \textbf{7.69\SPSB{\,\fontsize{5}{1}\selectfont +0.06}{\,\fontsize{5}{1}\selectfont -0.08}} & {\bfseries 100.0} & 100.0     & 1819.2    & 9.93 $\pm$ 0.13    & 27.2                  \\[1mm]
      Predictive 1        & 7.41\SPSB{\,\fontsize{5}{1}\selectfont +0.07}{\,\fontsize{5}{1}\selectfont -0.09} &  96.3 & 100.0                          & 1819.2    & 9.93 $\pm$ 0.13    & 27.2                  \\[1mm]
      Predictive 2        & 6.51\SPSB{\,\fontsize{5}{1}\selectfont +0.03}{\,\fontsize{5}{1}\selectfont -0.08} &  84.7 & 100.0                          & 1819.2    & 9.93 $\pm$ 0.13    & 27.2                  \\
      \midrule                      
      S-CC 1              & 7.15\SPSB{\,\fontsize{5}{1}\selectfont +0.12}{\,\fontsize{5}{1}\selectfont -0.13} &  93.0  & {\bfseries 50.1}                           & {\bfseries 911.4}    & \textbf{5.28 $\pm$ 0.07}    & {\bfseries 14.6}                      \\[1mm]
      S-CC 2              & 7.23\SPSB{\,\fontsize{5}{1}\selectfont +0.04}{\,\fontsize{5}{1}\selectfont -0.05} &  94.0  & 51.4                           & 935.2    & 5.63 $\pm$ 0.11    & 18.7                      \\[1mm]
      S-CC 3              & 7.28\SPSB{\,\fontsize{5}{1}\selectfont +0.07}{\,\fontsize{5}{1}\selectfont -0.05} &  94.7  & 58.1                           & 1057.5    & 6.27 $\pm$ 0.10    & 24.0                      \\[1mm]
      S-CC 4              & 7.43\SPSB{\,\fontsize{5}{1}\selectfont +0.18}{\,\fontsize{5}{1}\selectfont -0.14} &  96.7  & 61.5                           & 1118.3    & 6.67 $\pm$ 0.13    & 25.1                      \\[1mm]
      S-CC 5              & 7.47\SPSB{\,\fontsize{5}{1}\selectfont +0.07}{\,\fontsize{5}{1}\selectfont -0.15} &  97.2  & 64.8                           & 1178.7   & 6.98 $\pm$ 0.14    & 25.6                       \\[1mm]
      S-CC 6              & 7.56\SPSB{\,\fontsize{5}{1}\selectfont +0.05}{\,\fontsize{5}{1}\selectfont -0.06} &  98.3  & 71.3                           & 1296.9   & 7.50 $\pm$ 0.08    & 26.1                       \\[1mm]
      S-CC 7              & 7.55\SPSB{\,\fontsize{5}{1}\selectfont +0.05}{\,\fontsize{5}{1}\selectfont -0.05} &  98.2  & 83.8                           & 1524.3   & 8.43 $\pm$ 0.12    & 26.6                       \\
      \bottomrule
    \end{tabular}
    }
    \label{table:inferene_time_and_memory}
  \end{table}

\begin{figure}[H]
  \centering
  \includegraphics[width=0.9\textwidth]{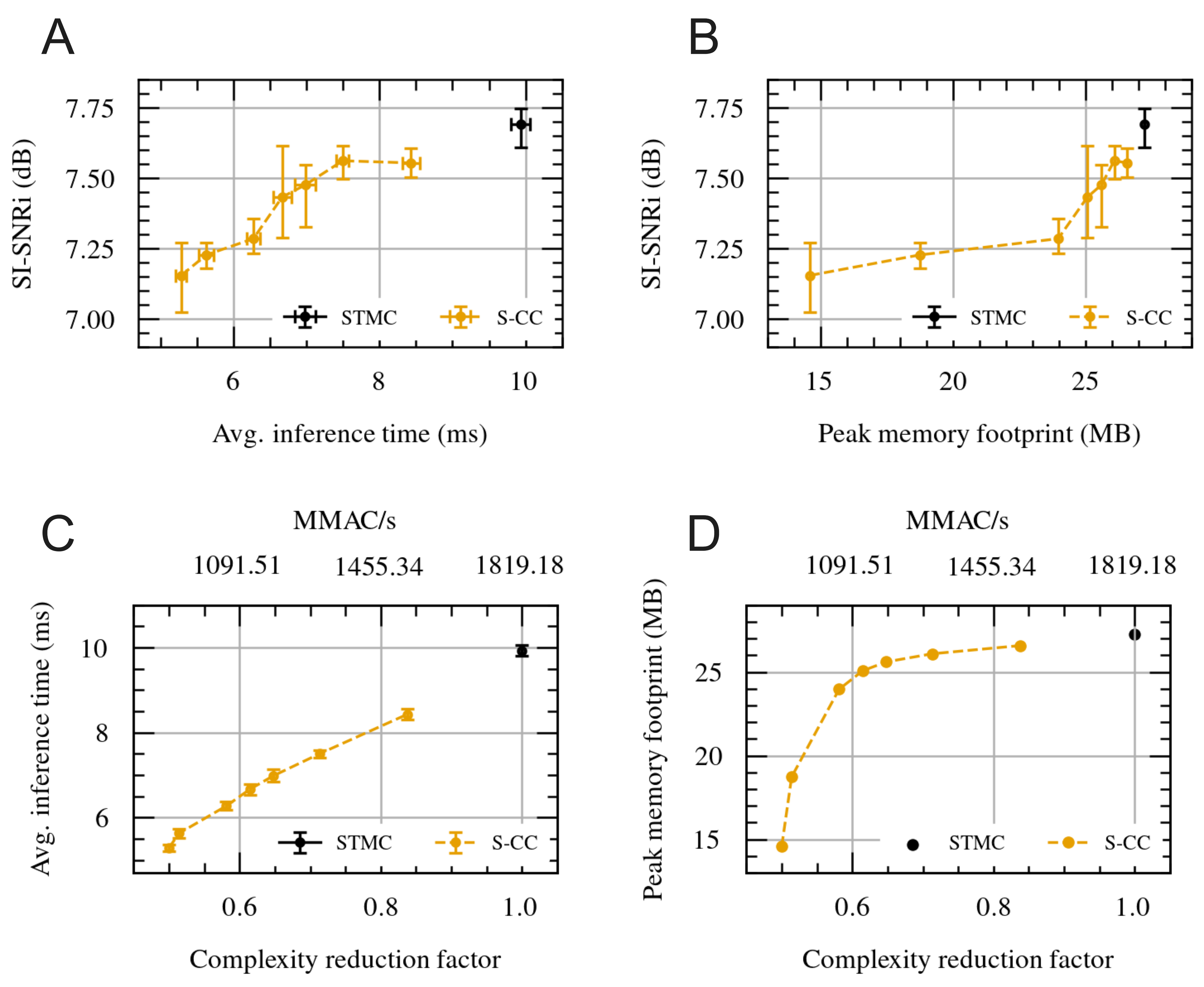}
  \caption{Average inference time and peak memory footprint }
  \label{fig:inferene_time_and_memory}
\end{figure}

Additionally, in Figure \ref{fig:inferene_time_and_memory}, we show how inference time and peak memory consumption depend on SI-SNRi and the complexity reduction factor.

These results suggest that the inference time of the tested model depends linearly on SI-SNRi and the complexity reduction factor, while peak memory consumption has significant impact on SI-SNRi at the beginning. From 60\% complexity reduction onward, the peak memory footprint decreases sharply.

\section{Interpolation}

\begin{wrapfigure}[14]{r}{7.5cm}
    \vspace{-1.3cm}
    \centering
    \includegraphics[width=7.2cm]{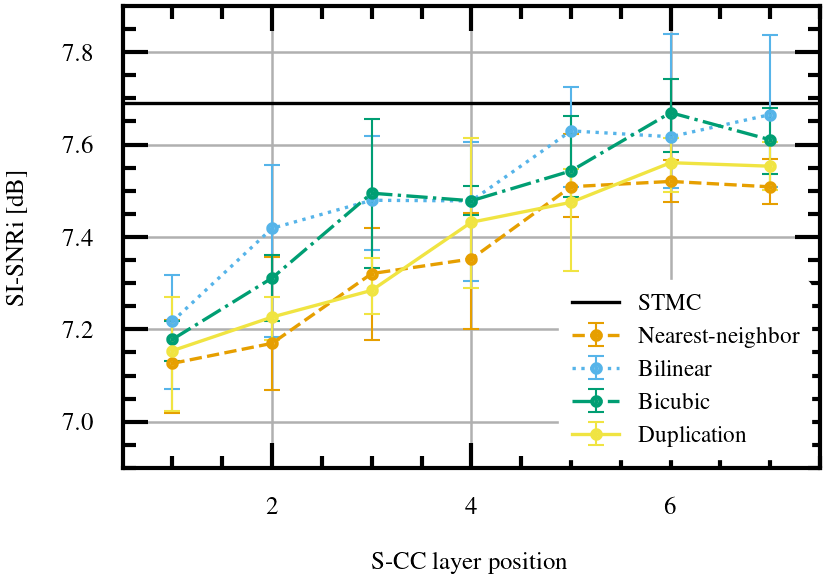}
    \caption{Comparison between extrapolated frame duplication and interpolation methods for PP SOI.}
    \label{fig:interpolation}
\end{wrapfigure}

Our method may also benefit from the usage of interpolation methods in place of extrapolation. We tested a singular S-CC layer with three different types of interpolation: nearest-neighbor, bilinear, and bicubic. The results of this experiment can be observed in Figure \ref{fig:interpolation} and Table \ref{table:results_interpolation}. For comparison, we included our extrapolated duplication method in the results of this experiment.

In this experiment, we achieved the best results with bilinear or bicubic interpolation, although the results for bilinear interpolation showed much higher variance than any other method. It is important to remember that even though we achieve slightly better results using interpolation, the usage of interpolation comes at the cost of higher latency as we need to wait for an additional time frame.

\begin{table}[H]
    \vspace{-0.25cm}
  \caption{Results of experiment on interpolation methods with PP SOI.}
  \centering
  {
  \setlength{\tabcolsep}{3pt}
  \begin{tabular}{l@{\hskip 5mm} c@{\hskip 5mm} c@{\hskip 5mm} c@{\hskip 5mm} c}
    \toprule
    \multirow{5}{*}{Model}               & \multicolumn{4}{c}{SI-SNRi (dB)}  \\
    \cmidrule(lr){2-5}
     & \multirow{2}{*}{Duplication} & Nearest- & \multirow{2}{*}{Bilinear} & \multirow{2}{*}{Bicubic}  \\
     &                              & neighbor & & \\
    \midrule
    S-CC 1 
    & 7.15\SPSB{\,\fontsize{6}{1}\selectfont +0.12}{\,\fontsize{6}{1}\selectfont -0.13}
    & 7.13\SPSB{\,\fontsize{6}{1}\selectfont +0.10}{\,\fontsize{6}{1}\selectfont -0.11}
    & \textbf{7.22}\SPSB{\,\fontsize{6}{1}\selectfont +0.11}{\,\fontsize{6}{1}\selectfont -0.15}
    & 7.18\SPSB{\,\fontsize{6}{1}\selectfont +0.05}{\,\fontsize{6}{1}\selectfont -0.05}
    \\[1mm]
    S-CC 2 
    & 7.23\SPSB{\,\fontsize{6}{1}\selectfont +0.19}{\,\fontsize{6}{1}\selectfont -0.11}
    & 7.17\SPSB{\,\fontsize{6}{1}\selectfont +0.06}{\,\fontsize{6}{1}\selectfont -0.05}
    & \textbf{7.42}\SPSB{\,\fontsize{6}{1}\selectfont +0.14}{\,\fontsize{6}{1}\selectfont -0.24}
    & 7.31\SPSB{\,\fontsize{6}{1}\selectfont +0.05}{\,\fontsize{6}{1}\selectfont -0.10}
    \\[1mm]
    S-CC 3 
    & 7.28\SPSB{\,\fontsize{6}{1}\selectfont +0.07}{\,\fontsize{6}{1}\selectfont -0.05}
    & 7.32\SPSB{\,\fontsize{6}{1}\selectfont +0.10}{\,\fontsize{6}{1}\selectfont -0.15}
    & 7.48\SPSB{\,\fontsize{6}{1}\selectfont +0.14}{\,\fontsize{6}{1}\selectfont -0.11}
    & \textbf{7.49}\SPSB{\,\fontsize{6}{1}\selectfont +0.17}{\,\fontsize{6}{1}\selectfont -0.17}
    \\[1mm]
    S-CC 4
    & 7.43\SPSB{\,\fontsize{6}{1}\selectfont +0.18}{\,\fontsize{6}{1}\selectfont -0.14}
    & 7.35\SPSB{\,\fontsize{6}{1}\selectfont +0.11}{\,\fontsize{6}{1}\selectfont -0.16}
    & \textbf{7.48}\SPSB{\,\fontsize{6}{1}\selectfont +0.13}{\,\fontsize{6}{1}\selectfont -0.18}
    & \textbf{7.48}\SPSB{\,\fontsize{6}{1}\selectfont +0.04}{\,\fontsize{6}{1}\selectfont -0.04}
    \\[1mm]
    S-CC 5 
    & 7.47\SPSB{\,\fontsize{6}{1}\selectfont +0.07}{\,\fontsize{6}{1}\selectfont -0.15}
    & 7.51\SPSB{\,\fontsize{6}{1}\selectfont +0.12}{\,\fontsize{6}{1}\selectfont -0.07}
    & \textbf{7.63}\SPSB{\,\fontsize{6}{1}\selectfont +0.10}{\,\fontsize{6}{1}\selectfont -0.13}
    & 7.54\SPSB{\,\fontsize{6}{1}\selectfont +0.12}{\,\fontsize{6}{1}\selectfont -0.06}
    \\[1mm]
    S-CC 6 
    & 7.56\SPSB{\,\fontsize{6}{1}\selectfont +0.05}{\,\fontsize{6}{1}\selectfont -0.06}
    & 7.52\SPSB{\,\fontsize{6}{1}\selectfont +0.05}{\,\fontsize{6}{1}\selectfont -0.05}
    & 7.62\SPSB{\,\fontsize{6}{1}\selectfont +0.23}{\,\fontsize{6}{1}\selectfont -0.12}
    & \textbf{7.67}\SPSB{\,\fontsize{6}{1}\selectfont +0.08}{\,\fontsize{6}{1}\selectfont -0.09}
    \\[1mm]
    S-CC 7 
    & 7.55\SPSB{\,\fontsize{6}{1}\selectfont +0.05}{\,\fontsize{6}{1}\selectfont -0.05}
    & 7.51\SPSB{\,\fontsize{6}{1}\selectfont +0.07}{\,\fontsize{6}{1}\selectfont -0.04}
    & \textbf{7.66}\SPSB{\,\fontsize{6}{1}\selectfont +0.18}{\,\fontsize{6}{1}\selectfont -0.16}
    & 7.61\SPSB{\,\fontsize{6}{1}\selectfont +0.07}{\,\fontsize{6}{1}\selectfont -0.08}
    \\
    \bottomrule
  \end{tabular}
  }
  \label{table:results_interpolation}
\end{table}

\section{Different Extrapolation Methods \label{section:extrapolation}}

In the main paper we used element duplication as an extrapolation method but we also pointed out that any type of extrapolation may be used. Here we compare the results of our speech separation experiment using element duplication with another commonly found method -- a transposed convolution. We tested both partially predictive and fully predictive modes using the setup described in Section \ref{section:SE}. 

\begin{figure}[h]
    \centering
    \ifdefined\enableColors
        \includegraphics[width=0.95\textwidth]{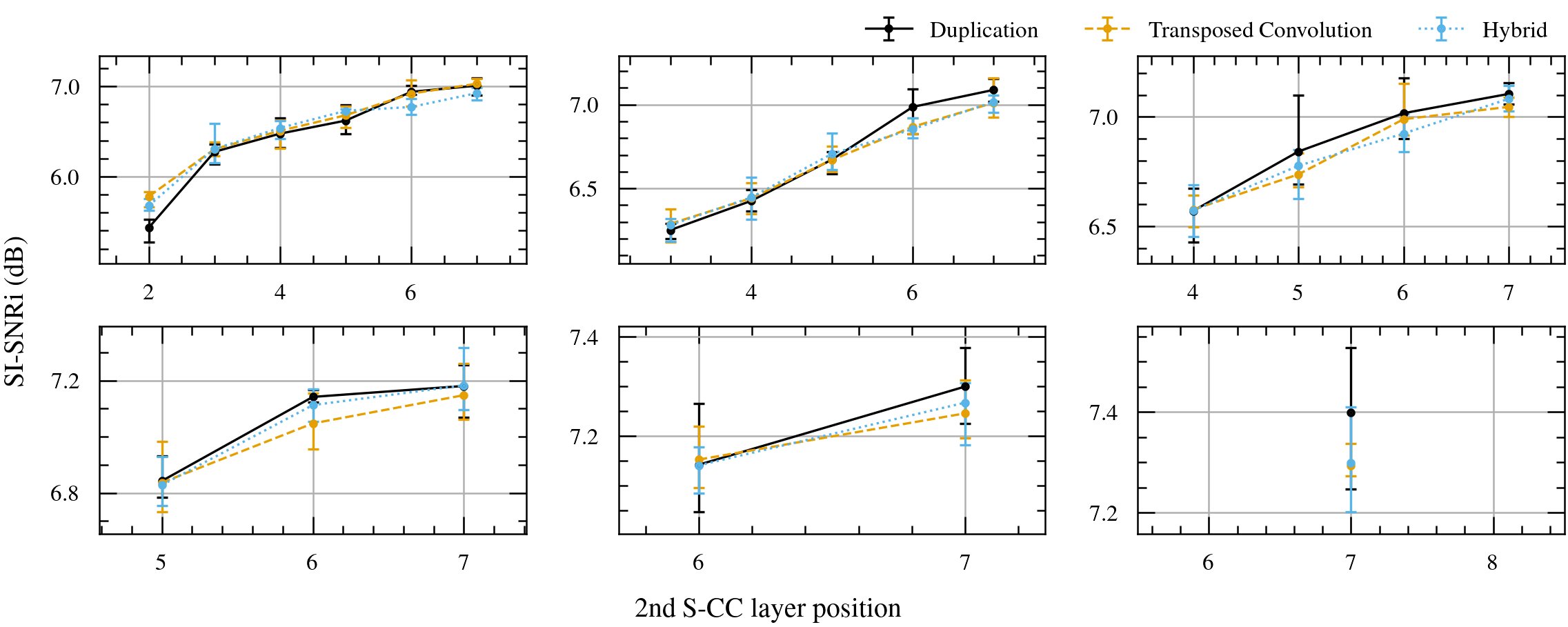}
    \else
        \includegraphics[width=0.95\textwidth]{images/appendix/2xS-CC_tconv_vs_copy.png}
    \fi
    \caption{Comparison between frame duplication and transposed convolution for PP SOI.}
    \label{fig:STC_vs_SOI_2xS-CC}
\end{figure}

Results of the experiment with PP SOI are presented in figure \ref{fig:STC_vs_SOI_2xS-CC}. In this case we tested a U-Net with two S-CC layers. Each plot represents a different position of the first S-CC pair and each point on X-axis represents a different position of the second S-CC pair. We also present results for hybrid models where duplication and transposed convolutions were used together -- in the first and second S-CC pairs respectively.

In this experiment neither method demonstrated a significant advantage. It seems that duplication tends to perform better than transposed convolution if it is introduced deeper within the network and \emph{vice versa} although difference is marginal.

\begin{figure}[h]
    \centering
    \ifdefined\enableColors
        \includegraphics[width=0.95\textwidth]{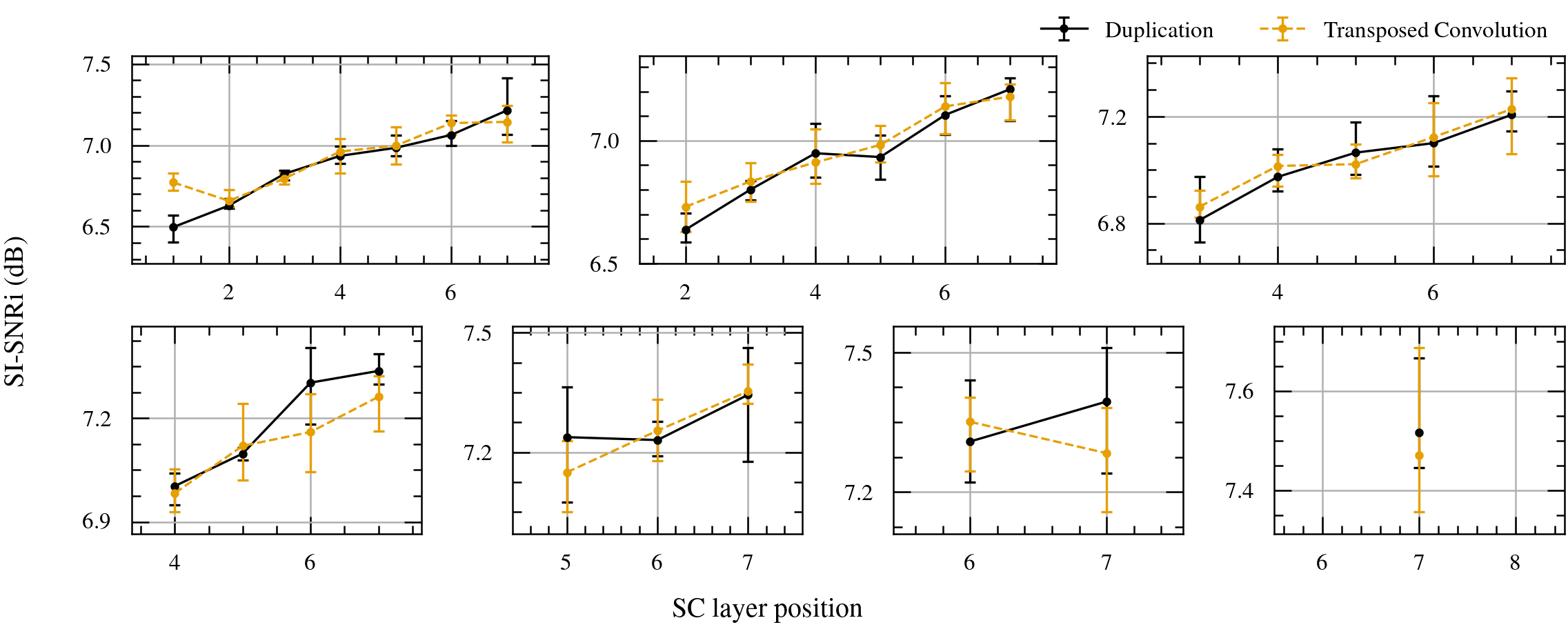}
    \else
        \includegraphics[width=0.95\textwidth]{images/appendix/FP_tconv_vs_copy.png}
    \fi
    \caption{Comparison between frame duplication and transposed convolution for FP SOI.}
    \label{fig:STC_vs_SOI_FP}
\end{figure}

Results achieved with FP SOI are shown in figure \ref{fig:STC_vs_SOI_FP}. Each plot represents a different position of S-CC pair and each point represents different position of SC layer. Achieved results confirm previous conclusions. 

This experiment proves that element duplication is a viable method and thus should be chosen for its simplicity.

\begin{table}[H]
    \vspace{-0.5cm}
  \caption{Results of experiment on different extrapolation methods with PP SOI.}
  \centering
  {
  \setlength{\tabcolsep}{3pt}
  \begin{tabular}{l@{\hskip 5mm} c@{\hskip 5mm} c@{\hskip 5mm} c}
    \toprule
    \multirow{3}{*}{Model}               & \multicolumn{3}{c}{SI-SNRi (dB)}  \\
    \cmidrule(lr){2-4}
     & \multirow{2}{*}{Duplication} & Transposed  & \multirow{2}{*}{Hybrid} \\
     &                              & convolution &                         \\
    \midrule
    S-CC 1$\vert$2 
    & 5.43\SPSB{\,\fontsize{6}{1}\selectfont +0.07}{\,\fontsize{6}{1}\selectfont -0.17}
    & \textbf{5.78}\SPSB{\,\fontsize{6}{1}\selectfont +0.05}{\,\fontsize{6}{1}\selectfont -0.12}
    & 5.68\SPSB{\,\fontsize{6}{1}\selectfont +0.08}{\,\fontsize{6}{1}\selectfont -0.06}
    \\[1mm]
    S-CC 1$\vert$3 
    & 6.27\SPSB{\,\fontsize{6}{1}\selectfont +0.18}{\,\fontsize{6}{1}\selectfont -0.14}
    & 6.30\SPSB{\,\fontsize{6}{1}\selectfont +0.09}{\,\fontsize{6}{1}\selectfont -0.08}
    & \textbf{6.31}\SPSB{\,\fontsize{6}{1}\selectfont +0.29}{\,\fontsize{6}{1}\selectfont -0.16}
    \\[1mm]
    S-CC 1$\vert$4
    & 6.48\SPSB{\,\fontsize{6}{1}\selectfont +0.18}{\,\fontsize{6}{1}\selectfont -0.16}
    & 6.51\SPSB{\,\fontsize{6}{1}\selectfont +0.13}{\,\fontsize{6}{1}\selectfont -0.20}
    & \textbf{6.54}\SPSB{\,\fontsize{6}{1}\selectfont +0.08}{\,\fontsize{6}{1}\selectfont -0.11}
    \\[1mm]
    S-CC 1$\vert$5 
    & 6.62\SPSB{\,\fontsize{6}{1}\selectfont +0.09}{\,\fontsize{6}{1}\selectfont -0.15}
    & 6.68\SPSB{\,\fontsize{6}{1}\selectfont +0.11}{\,\fontsize{6}{1}\selectfont -0.14}
    & \textbf{6.73}\SPSB{\,\fontsize{6}{1}\selectfont +0.03}{\,\fontsize{6}{1}\selectfont -0.03}
    \\[1mm]
    S-CC 1$\vert$6 
    & \textbf{6.94}\SPSB{\,\fontsize{6}{1}\selectfont +0.10}{\,\fontsize{6}{1}\selectfont -0.04}
    & 6.92\SPSB{\,\fontsize{6}{1}\selectfont +0.15}{\,\fontsize{6}{1}\selectfont -0.14}
    & 6.77\SPSB{\,\fontsize{6}{1}\selectfont +0.10}{\,\fontsize{6}{1}\selectfont -0.09}
    \\[1mm]
    S-CC 1$\vert$7 
    & 7.01\SPSB{\,\fontsize{6}{1}\selectfont +0.09}{\,\fontsize{6}{1}\selectfont -0.11}
    & \textbf{7.03}\SPSB{\,\fontsize{6}{1}\selectfont +0.07}{\,\fontsize{6}{1}\selectfont -0.03}
    & 6.92\SPSB{\,\fontsize{6}{1}\selectfont +0.07}{\,\fontsize{6}{1}\selectfont -0.08}
    \\
    \midrule
    S-CC 2$\vert$3
    & 6.25\SPSB{\,\fontsize{6}{1}\selectfont +0.04}{\,\fontsize{6}{1}\selectfont -0.06}
    & \textbf{6.29}\SPSB{\,\fontsize{6}{1}\selectfont +0.09}{\,\fontsize{6}{1}\selectfont -0.11}
    & 6.28\SPSB{\,\fontsize{6}{1}\selectfont +0.04}{\,\fontsize{6}{1}\selectfont -0.10}
    \\[1mm]
    S-CC 2$\vert$4
    & 6.43\SPSB{\,\fontsize{6}{1}\selectfont +0.07}{\,\fontsize{6}{1}\selectfont -0.07}
    & 6.44\SPSB{\,\fontsize{6}{1}\selectfont +0.10}{\,\fontsize{6}{1}\selectfont -0.10}
    & \textbf{6.45}\SPSB{\,\fontsize{6}{1}\selectfont +0.12}{\,\fontsize{6}{1}\selectfont -0.13}
    \\[1mm]
    S-CC 2$\vert$5
    & 6.67\SPSB{\,\fontsize{6}{1}\selectfont +0.05}{\,\fontsize{6}{1}\selectfont -0.09}
    & 6.67\SPSB{\,\fontsize{6}{1}\selectfont +0.09}{\,\fontsize{6}{1}\selectfont -0.07}
    & \textbf{6.71}\SPSB{\,\fontsize{6}{1}\selectfont +0.13}{\,\fontsize{6}{1}\selectfont -0.10}
    \\[1mm]
    S-CC 2$\vert$6
    & \textbf{6.99}\SPSB{\,\fontsize{6}{1}\selectfont +0.11}{\,\fontsize{6}{1}\selectfont -0.16}
    & 6.87\SPSB{\,\fontsize{6}{1}\selectfont +0.06}{\,\fontsize{6}{1}\selectfont -0.05}
    & 6.85\SPSB{\,\fontsize{6}{1}\selectfont +0.07}{\,\fontsize{6}{1}\selectfont -0.06}
    \\[1mm]
    S-CC 2$\vert$7 
    & \textbf{7.09}\SPSB{\,\fontsize{6}{1}\selectfont +0.07}{\,\fontsize{6}{1}\selectfont -0.07}
    & 7.01\SPSB{\,\fontsize{6}{1}\selectfont +0.15}{\,\fontsize{6}{1}\selectfont -0.09}
    & 7.02\SPSB{\,\fontsize{6}{1}\selectfont +0.05}{\,\fontsize{6}{1}\selectfont -0.07}
    \\
    \midrule
    S-CC 3$\vert$4
    & 6.57\SPSB{\,\fontsize{6}{1}\selectfont +0.11}{\,\fontsize{6}{1}\selectfont -0.15}
    & \textbf{6.58}\SPSB{\,\fontsize{6}{1}\selectfont +0.07}{\,\fontsize{6}{1}\selectfont -0.08}
    & 6.57\SPSB{\,\fontsize{6}{1}\selectfont +0.12}{\,\fontsize{6}{1}\selectfont -0.12}
    \\[1mm]    
    S-CC 3$\vert$5
    & \textbf{6.84}\SPSB{\,\fontsize{6}{1}\selectfont +0.26}{\,\fontsize{6}{1}\selectfont -0.15}
    & 6.74\SPSB{\,\fontsize{6}{1}\selectfont +0.10}{\,\fontsize{6}{1}\selectfont -0.06}
    & 6.78\SPSB{\,\fontsize{6}{1}\selectfont +0.08}{\,\fontsize{6}{1}\selectfont -0.15}
    \\[1mm]
    S-CC 3$\vert$6 
    & \textbf{7.02}\SPSB{\,\fontsize{6}{1}\selectfont +0.17}{\,\fontsize{6}{1}\selectfont -0.12}
    & 6.99\SPSB{\,\fontsize{6}{1}\selectfont +0.17}{\,\fontsize{6}{1}\selectfont -0.08}
    & 6.92\SPSB{\,\fontsize{6}{1}\selectfont +0.09}{\,\fontsize{6}{1}\selectfont -0.09}
    \\[1mm]
    S-CC 3$\vert$7 
    & \textbf{7.10}\SPSB{\,\fontsize{6}{1}\selectfont +0.05}{\,\fontsize{6}{1}\selectfont -0.05}
    & 7.05\SPSB{\,\fontsize{6}{1}\selectfont +0.05}{\,\fontsize{6}{1}\selectfont -0.05}
    & 7.08\SPSB{\,\fontsize{6}{1}\selectfont +0.07}{\,\fontsize{6}{1}\selectfont -0.06}
    \\
    \midrule
    S-CC 4$\vert$5
    & \textbf{6.84}\SPSB{\,\fontsize{6}{1}\selectfont +0.09}{\,\fontsize{6}{1}\selectfont -0.06}
    & \textbf{6.84}\SPSB{\,\fontsize{6}{1}\selectfont +0.15}{\,\fontsize{6}{1}\selectfont -0.11}
    & 6.83\SPSB{\,\fontsize{6}{1}\selectfont +0.11}{\,\fontsize{6}{1}\selectfont -0.08}
    \\[1mm] 
    S-CC 4$\vert$6
    & \textbf{7.14}\SPSB{\,\fontsize{6}{1}\selectfont +0.03}{\,\fontsize{6}{1}\selectfont -0.18}
    & 7.05\SPSB{\,\fontsize{6}{1}\selectfont +0.11}{\,\fontsize{6}{1}\selectfont -0.10}
    & 7.11\SPSB{\,\fontsize{6}{1}\selectfont +0.06}{\,\fontsize{6}{1}\selectfont -0.06}
    \\[1mm]    
    S-CC 4$\vert$7
    & \textbf{7.18}\SPSB{\,\fontsize{6}{1}\selectfont +0.08}{\,\fontsize{6}{1}\selectfont -0.12}
    & 7.15\SPSB{\,\fontsize{6}{1}\selectfont +0.12}{\,\fontsize{6}{1}\selectfont -0.09}
    & \textbf{7.18}\SPSB{\,\fontsize{6}{1}\selectfont +0.14}{\,\fontsize{6}{1}\selectfont -0.09}
    \\   
    \midrule
    S-CC 5$\vert$6
    & 7.14\SPSB{\,\fontsize{6}{1}\selectfont +0.13}{\,\fontsize{6}{1}\selectfont -0.10}
    & \textbf{7.15}\SPSB{\,\fontsize{6}{1}\selectfont +0.07}{\,\fontsize{6}{1}\selectfont -0.06}
    & 7.14\SPSB{\,\fontsize{6}{1}\selectfont +0.04}{\,\fontsize{6}{1}\selectfont -0.06}
    \\[1mm]   
    S-CC 5$\vert$7
    & \textbf{7.30}\SPSB{\,\fontsize{6}{1}\selectfont +0.08}{\,\fontsize{6}{1}\selectfont -0.08}
    & 7.25\SPSB{\,\fontsize{6}{1}\selectfont +0.07}{\,\fontsize{6}{1}\selectfont -0.06}
    & 7.27\SPSB{\,\fontsize{6}{1}\selectfont +0.04}{\,\fontsize{6}{1}\selectfont -0.09}
    \\   
    \midrule
    S-CC 6$\vert$7
    & \textbf{7.34}\SPSB{\,\fontsize{6}{1}\selectfont +0.13}{\,\fontsize{6}{1}\selectfont -0.16}
    & 7.29\SPSB{\,\fontsize{6}{1}\selectfont +0.05}{\,\fontsize{6}{1}\selectfont -0.03}
    & 7.30\SPSB{\,\fontsize{6}{1}\selectfont +0.12}{\,\fontsize{6}{1}\selectfont -0.10}
    \\  
    \bottomrule
  \end{tabular}
  }
  \label{table:results_PP_app_B}
\end{table}

\begin{table}[H]
\vspace{-3mm}
  \caption{Results of experiment on different extrapolation methods with FP SOI.}
  \centering
  {
  \setlength{\tabcolsep}{3pt}
  \begin{tabular}{l@{\hskip 5mm} c@{\hskip 5mm} c}
    \toprule
    \multirow{3}{*}{Model}               & \multicolumn{2}{c}{SI-SNRi (dB)}  \\
    \cmidrule(lr){2-3}
     & \multirow{2}{*}{Duplication} & Transposed  \\
     &                              & convolution \\
    \midrule
    SS-CC 1 
    & 6.50\SPSB{\,\fontsize{6}{1}\selectfont +0.08}{\,\fontsize{6}{1}\selectfont -0.10}
    & \textbf{6.77}\SPSB{\,\fontsize{6}{1}\selectfont +0.06}{\,\fontsize{6}{1}\selectfont -0.05}
    \\[1mm]
    S-CC 1$\vert$2 
    & 6.63\SPSB{\,\fontsize{6}{1}\selectfont +0.05}{\,\fontsize{6}{1}\selectfont -0.02}
    & \textbf{6.66}\SPSB{\,\fontsize{6}{1}\selectfont +0.07}{\,\fontsize{6}{1}\selectfont -0.04}
    \\[1mm]
    S-CC 1$\vert$3 
    & \textbf{6.82}\SPSB{\,\fontsize{6}{1}\selectfont +0.03}{\,\fontsize{6}{1}\selectfont -0.04}
    & 6.80\SPSB{\,\fontsize{6}{1}\selectfont +0.04}{\,\fontsize{6}{1}\selectfont -0.04}
    \\[1mm]
    S-CC 1$\vert$4
    & 6.93\SPSB{\,\fontsize{6}{1}\selectfont +0.07}{\,\fontsize{6}{1}\selectfont -0.05}
    & \textbf{6.96}\SPSB{\,\fontsize{6}{1}\selectfont +0.08}{\,\fontsize{6}{1}\selectfont -0.14}
    \\[1mm]
    S-CC 1$\vert$5 
    & 6.98\SPSB{\,\fontsize{6}{1}\selectfont +0.08}{\,\fontsize{6}{1}\selectfont -0.06}
    & \textbf{7.00}\SPSB{\,\fontsize{6}{1}\selectfont +0.12}{\,\fontsize{6}{1}\selectfont -0.12}
    \\[1mm]
    S-CC 1$\vert$6 
    & 7.06\SPSB{\,\fontsize{6}{1}\selectfont +0.09}{\,\fontsize{6}{1}\selectfont -0.07}
    & \textbf{7.14}\SPSB{\,\fontsize{6}{1}\selectfont +0.05}{\,\fontsize{6}{1}\selectfont -0.08}
    \\[1mm]
    S-CC 1$\vert$7 
    & \textbf{7.21}\SPSB{\,\fontsize{6}{1}\selectfont +0.20}{\,\fontsize{6}{1}\selectfont -0.15}
    & 7.14\SPSB{\,\fontsize{6}{1}\selectfont +0.11}{\,\fontsize{6}{1}\selectfont -0.14}
    \\
    \midrule
    SS-CC 2
    & 6.64\SPSB{\,\fontsize{6}{1}\selectfont +0.05}{\,\fontsize{6}{1}\selectfont -0.13}
    & \textbf{6.73}\SPSB{\,\fontsize{6}{1}\selectfont +0.11}{\,\fontsize{6}{1}\selectfont -0.11}
    \\[1mm]
    S-CC 2$\vert$3
    & 6.80\SPSB{\,\fontsize{6}{1}\selectfont +0.08}{\,\fontsize{6}{1}\selectfont -0.08}
    & \textbf{6.83}\SPSB{\,\fontsize{6}{1}\selectfont +0.08}{\,\fontsize{6}{1}\selectfont -0.09}
    \\[1mm]
    S-CC 2$\vert$4
    & \textbf{6.95}\SPSB{\,\fontsize{6}{1}\selectfont +0.10}{\,\fontsize{6}{1}\selectfont -0.10}
    & 6.91\SPSB{\,\fontsize{6}{1}\selectfont +0.14}{\,\fontsize{6}{1}\selectfont -0.09}
    \\[1mm]
    S-CC 2$\vert$5
    & 6.93\SPSB{\,\fontsize{6}{1}\selectfont +0.13}{\,\fontsize{6}{1}\selectfont -0.10}
    & \textbf{6.98}\SPSB{\,\fontsize{6}{1}\selectfont +0.08}{\,\fontsize{6}{1}\selectfont -0.07}
    \\[1mm]
    S-CC 2$\vert$6
    & 7.10\SPSB{\,\fontsize{6}{1}\selectfont +0.04}{\,\fontsize{6}{1}\selectfont -0.05}
    & \textbf{7.14}\SPSB{\,\fontsize{6}{1}\selectfont +0.10}{\,\fontsize{6}{1}\selectfont -0.12}
    \\[1mm]
    S-CC 2$\vert$7 
    & \textbf{7.21}\SPSB{\,\fontsize{6}{1}\selectfont +0.07}{\,\fontsize{6}{1}\selectfont -0.06}
    & 7.18\SPSB{\,\fontsize{6}{1}\selectfont +0.06}{\,\fontsize{6}{1}\selectfont -0.10}
    \\
    \midrule
    SS-CC 3
    & 6.81\SPSB{\,\fontsize{6}{1}\selectfont +0.17}{\,\fontsize{6}{1}\selectfont -0.09}
    & \textbf{6.86}\SPSB{\,\fontsize{6}{1}\selectfont +0.07}{\,\fontsize{6}{1}\selectfont -0.04}
    \\[1mm] 
    S-CC 3$\vert$4
    & 6.97\SPSB{\,\fontsize{6}{1}\selectfont +0.11}{\,\fontsize{6}{1}\selectfont -0.06}
    & \textbf{7.01}\SPSB{\,\fontsize{6}{1}\selectfont +0.05}{\,\fontsize{6}{1}\selectfont -0.08}
    \\[1mm]    
    S-CC 3$\vert$5
    & \textbf{7.06}\SPSB{\,\fontsize{6}{1}\selectfont +0.12}{\,\fontsize{6}{1}\selectfont -0.09}
    & 7.02\SPSB{\,\fontsize{6}{1}\selectfont +0.08}{\,\fontsize{6}{1}\selectfont -0.06}
    \\[1mm]
    S-CC 3$\vert$6 
    & 7.10\SPSB{\,\fontsize{6}{1}\selectfont +0.18}{\,\fontsize{6}{1}\selectfont -0.09}
    & \textbf{7.12}\SPSB{\,\fontsize{6}{1}\selectfont +0.13}{\,\fontsize{6}{1}\selectfont -0.15}
    \\[1mm]
    S-CC 3$\vert$7 
    & 7.21\SPSB{\,\fontsize{6}{1}\selectfont +0.09}{\,\fontsize{6}{1}\selectfont -0.07}
    & \textbf{7.23}\SPSB{\,\fontsize{6}{1}\selectfont +0.12}{\,\fontsize{6}{1}\selectfont -0.17}
    \\
    \midrule
    SS-CC 4
    & \textbf{7.00}\SPSB{\,\fontsize{6}{1}\selectfont +0.04}{\,\fontsize{6}{1}\selectfont -0.06}
    & 6.98\SPSB{\,\fontsize{6}{1}\selectfont +0.08}{\,\fontsize{6}{1}\selectfont -0.06}
    \\[1mm] 
    S-CC 4$\vert$5
    & 7.10\SPSB{\,\fontsize{6}{1}\selectfont +0.02}{\,\fontsize{6}{1}\selectfont -0.02}
    & \textbf{7.12}\SPSB{\,\fontsize{6}{1}\selectfont +0.13}{\,\fontsize{6}{1}\selectfont -0.10}
    \\[1mm] 
    S-CC 4$\vert$6
    & \textbf{7.30}\SPSB{\,\fontsize{6}{1}\selectfont +0.11}{\,\fontsize{6}{1}\selectfont -0.13}
    & 7.16\SPSB{\,\fontsize{6}{1}\selectfont +0.12}{\,\fontsize{6}{1}\selectfont -0.12}
    \\[1mm]    
    S-CC 4$\vert$7
    & \textbf{7.34}\SPSB{\,\fontsize{6}{1}\selectfont +0.05}{\,\fontsize{6}{1}\selectfont -0.04}
    & 7.26\SPSB{\,\fontsize{6}{1}\selectfont +0.07}{\,\fontsize{6}{1}\selectfont -0.10}
    \\   
    \midrule
    SS-CC 5
    & \textbf{7.24}\SPSB{\,\fontsize{6}{1}\selectfont +0.13}{\,\fontsize{6}{1}\selectfont -0.17}
    & 7.15\SPSB{\,\fontsize{6}{1}\selectfont +0.08}{\,\fontsize{6}{1}\selectfont -0.10}
    \\[1mm]   
    S-CC 5$\vert$6
    & 7.23\SPSB{\,\fontsize{6}{1}\selectfont +0.05}{\,\fontsize{6}{1}\selectfont -0.04}
    & \textbf{7.25}\SPSB{\,\fontsize{6}{1}\selectfont +0.08}{\,\fontsize{6}{1}\selectfont -0.08}
    \\[1mm]   
    S-CC 5$\vert$7
    & 7.34\SPSB{\,\fontsize{6}{1}\selectfont +0.12}{\,\fontsize{6}{1}\selectfont -0.17}
    & \textbf{7.35}\SPSB{\,\fontsize{6}{1}\selectfont +0.07}{\,\fontsize{6}{1}\selectfont -0.04}
    \\
    \midrule
    SS-CC 6
    & 7.31\SPSB{\,\fontsize{6}{1}\selectfont +0.14}{\,\fontsize{6}{1}\selectfont -0.09}
    & \textbf{7.35}\SPSB{\,\fontsize{6}{1}\selectfont +0.06}{\,\fontsize{6}{1}\selectfont -0.11}
    \\[1mm]   
    S-CC 6$\vert$7
    & \textbf{7.39}\SPSB{\,\fontsize{6}{1}\selectfont +0.12}{\,\fontsize{6}{1}\selectfont -0.16}
    & 7.28\SPSB{\,\fontsize{6}{1}\selectfont +0.10}{\,\fontsize{6}{1}\selectfont -0.13}
    \\  
    \midrule
    SS-CC 7
    & \textbf{7.52}\SPSB{\,\fontsize{6}{1}\selectfont +0.16}{\,\fontsize{6}{1}\selectfont -0.08}
    & 7.47\SPSB{\,\fontsize{6}{1}\selectfont +0.22}{\,\fontsize{6}{1}\selectfont -0.12}
    \\  
    \bottomrule
  \end{tabular}
  }
  \label{table:results_FP_app_B}
\end{table}

\newpage

\section{Video Action Recognition}

Other domains can benefit from SOI, as it can be applied to any time-series data. To illustrate this, we conducted experiments using SOI for an action recognition task with video data. We utilized the HMDB-51 dataset \cite{Kuehne11}, which contains 24 fps video data of human actions split into 51 classes. We trained a popular ResNet-10 architecture \cite{ResNet10} in three variants: regular, small (with halved number of channels) and tiny (with number of channels reduced fourfold), where we replaced 3D convolutional layers with 3D STMC layers. Here, we applied SOI by introducing a skip connection between the output of block 2 and the input of block 4, optimizing block 3.

\begin{table}[H]
    \vspace{-0.25cm}
  \caption{Results of video action recognition experiment.}
  \centering
  {
  \setlength{\tabcolsep}{3pt}
  \begin{tabular}{l@{\hskip 8mm} c@{\hskip 5mm} c  c @{\hskip 8mm} c @{\hskip 5mm} c}
    \toprule
    \multirow{3}{*}{Model}               & \multicolumn{2}{c}{Regular} & & \multicolumn{2}{c}{SOI} \\
    \cmidrule(lr){2-3}\cmidrule(lr){5-6}
     & Top-1 Acc  & Complexity & & Top-1 Acc & Complexity \\
     & (\%) & (GMAC/s) && (\%) & (GMAC/s) \\
    \midrule
    ResNet-10
    & 32.63
    & 48.54
    &
    & 33.34
    & 40.69
    \\[1mm]
    ResNet-10 small
    & 31.24
    & 15.05
    &
    & 31.41
    & 13.09
    \\[1mm]
    ResNet-10 tiny 
    & 30.46
    & 5.23
    &
    & 30.90
    & \bfseries{4.73}
    \\
    \midrule
    MoViNet A0
    & 34.40
    & 33.15
    &
    & 31.88
    & 24.26
    \\[1mm]
    MoViNet A1
    & \bfseries{35.96}
    & 69.77
    &
    & 32.73
    & 53.92
    \\  
    \bottomrule
  \end{tabular}
  }
  \label{table:results_video}
\end{table}

To demonstrate that SOI can work not only with STMC, we also applied it to MoViNets, which use a method called ``Stream buffers''. We trained two variants of MoViNets, A0 and A1, in their streaming form. SOI was applied by optimizing blocks 4 and 5. Please note that SOI can be used not only with 3D convolutions but also with their popular 2D+1 variant.

All models were trained for 100 epochs with Adam optimizer and learning rate 5e-5. Each model was trained on Nvidia A100 GPU with batch size of 16. The mean training time amounted to about about 37h. 

Achieved results suggest that SOI is suitable for video domain as well. ResNet-10 architecture proved to be highly compatible with SOI for this task as results achieved with SOI variant of this model outperformed the regular ones. We believe that this improvement was imposed by the increase in receptive field as SOI adds additional strided convolution.  The achieved reduction of complexity for this family of models was between 10-17\% depending on model size. For MoViNets, the decrease of around 3\% in accuracy can be spotted, although the imposed complexity reduction was higher compared to ResNet and amounted to 23-30\%. 

\section{Acoustic Scene Classification with ResNet}

To further evaluate the effectiveness of our method, we conducted experiments on the accoustic scene classification task using ResNet architecture \cite{He2015DeepRL}. We chose ResNet for two primary reasons: (1) its widespread use and recognition as a standard deep learning model, and (2) its relatively large size compared to other architectures tested, such as GhostNet, which allows us to better assess how SOI performs on larger models.

We tested 4 different ResNet models for all 3 variants - Baseline, STMC and SOI. Each test was repeated 5 times. We used the TAU Urban Acoustic Scene 2020 Mobile dataset \citep{heittola_toni_2020_3819968} for both training and validation.

The results are presented in Table \ref{table:ASC_resnet_results}.

\begin{table}[H]
\caption{Results of ASC experiment with ResNet.}
\centering
{
\setlength{\tabcolsep}{3pt}

\begin{tabular}{c@{\hskip 5mm} c@{\hskip 5mm} c@{\hskip 5mm} c@{\hskip 5mm} c}
\toprule
\multirow{2}{*}{Model} & \multirow{2}{*}{Method} & Top-1         & Complexity & Number of  \\
                       &                         & Accuracy (\%) & (GMAC/s)   & Parameters \\
\midrule
\multirow{3.5}{*}{18}   & Baseline &         85.13\SPSB{\,\fontsize{6}{1}\selectfont +0.83}{\,\fontsize{6}{1}\selectfont -1.32}  &       143.65   &  \multirow{3.5}{*}{11.7M} \\[1mm]   
                     & STMC     &         85.13\SPSB{\,\fontsize{6}{1}\selectfont +0.83}{\,\fontsize{6}{1}\selectfont -1.32}  &         15.56   &   \\[1mm]   
                     & SOI      & \textbf{91.55\SPSB{\,\fontsize{6}{1}\selectfont +0.72}{\,\fontsize{6}{1}\selectfont -0.86}} & \textbf{12.35}  &            \\
\midrule
\multirow{3.5}{*}{34}  & Baseline & 86.03\SPSB{\,\fontsize{6}{1}\selectfont +1.36}{\,\fontsize{6}{1}\selectfont -3.08}  &       686.96   &  \multirow{3.5}{*}{21.8M} \\[1mm]   
                     & STMC     & 86.03\SPSB{\,\fontsize{6}{1}\selectfont +1.36}{\,\fontsize{6}{1}\selectfont -3.08}  &         32.65   &   \\[1mm]   
                     & SOI      &         \textbf{92.01\SPSB{\,\fontsize{6}{1}\selectfont +0.69}{\,\fontsize{6}{1}\selectfont -0.89}}   & \textbf{26.46}  &       \\
\midrule
\multirow{3.5}{*}{50} & Baseline &         89.66\SPSB{\,\fontsize{6}{1}\selectfont +0.60}{\,\fontsize{6}{1}\selectfont -0.54}  &      794.34   &  \multirow{3.5}{*}{25.6M} \\[1mm]   
                     & STMC     &         89.66\SPSB{\,\fontsize{6}{1}\selectfont +0.60}{\,\fontsize{6}{1}\selectfont -0.54}  &         33.10   &  \\[1mm]   
                     & SOI      & \textbf{91.43\SPSB{\,\fontsize{6}{1}\selectfont +0.83}{\,\fontsize{6}{1}\selectfont -1.89}} & \textbf{27.99}  &            \\
\midrule
\multirow{3.5}{*}{101}  & Baseline & 94.74\SPSB{\,\fontsize{6}{1}\selectfont +0.40}{\,\fontsize{6}{1}\selectfont -0.89}  &      2168.81       & \multirow{3.5}{*}{44.5M} \\[1mm]   
                     & STMC     & 94.74\SPSB{\,\fontsize{6}{1}\selectfont +0.40}{\,\fontsize{6}{1}\selectfont -0.89}  &         112.84   &  \\[1mm]   
                     & SOI      &         \textbf{96.22\SPSB{\,\fontsize{6}{1}\selectfont +0.92}{\,\fontsize{6}{1}\selectfont -0.95}}   & \textbf{95.83}  &     \\

\bottomrule
\end{tabular}
}
\label{table:ASC_resnet_results}
\end{table}

The results, as shown in Table \ref{table:ASC_resnet_results}, demonstrate that the SOI-enhanced ResNets consistently outperformed the baseline models in terms of accuracy. Specifically, the SOI-based models achieved higher classification accuracy across all tested configurations. We belived this improvement can be attributed to two key factors:

\begin{itemize}
    \item The introduction of SOI allows the model to generalize over longer time frames by expanding its receptive field. This helps the model capture more temporal context, which is crucial for tasks like ASC.
    \item SOI encourages the model to generalize its output states by predicting future partial states, which improves the model's ability to handle variations in the data.
\end{itemize}

Despite ResNet's depth and complexity, the method proved successful, not only reducing computational cost but also improving performance.

\end{document}